\pdfoutput=1

\documentclass[11pt]{article}

\usepackage[preprint]{acl}

\usepackage{times}
\usepackage{latexsym}

\usepackage[T1]{fontenc}

\usepackage[utf8]{inputenc}

\usepackage{microtype}

\usepackage{inconsolata}

\usepackage{graphicx}

\usepackage{makecell}
\usepackage{booktabs}

\usepackage{CJK}
\newcommand{\chinese}[1]{\begin{CJK}{UTF8}{gbsn}#1\end{CJK}}
\newcommand{\japanese}[1]{\begin{CJK}{UTF8}{min}#1\end{CJK}}
\newcommand{\korean}[1]{\begin{CJK}{UTF8}{mj}#1\end{CJK}}

\CJKnospace

\usepackage{multirow}

\usepackage{booktabs}

\title{Creative and Context-Aware Translation of East Asian Idioms with GPT-4}

\author{
  \textbf{Kenan Tang\textsuperscript{1 *}},
  \textbf{Peiyang Song\textsuperscript{2 *}},
  \textbf{Yao Qin\textsuperscript{1}},
  \textbf{Xifeng Yan\textsuperscript{1}}
  \\
  \textsuperscript{1 }UC Santa Barbara,
  \textsuperscript{2 }California Institute of Technology
  \\
  \small{
    \texttt{kenantang@ucsb.edu, psong@caltech.edu, yaoqin@ucsb.edu, xyan@cs.ucsb.edu}
  }
}

\definecolor{WildStrawberry}{RGB}{255,48,94}

\begin{document}
\maketitle
\begin{abstract}

As a type of figurative language, an East Asian idiom condenses rich cultural background into only a few characters. Translating such idioms is challenging for human translators, who often resort to choosing a context-aware translation from an existing list of candidates. However, compiling a dictionary of candidate translations demands much time and creativity even for expert translators. To alleviate such burden, we evaluate if GPT-4 can help generate high-quality translations. Based on automatic evaluations of faithfulness and creativity, we first identify Pareto-optimal prompting strategies that can outperform translation engines from Google and DeepL. Then, at a low cost, our context-aware translations can achieve far more high-quality translations per idiom than the human baseline. We open-source all code and data to facilitate further research\footnote{\url{https://github.com/kenantang/cjk-idioms-gpt}}.

\end{abstract}

\def\thefootnote{*}\footnotetext{Equal contributions.}\def\thefootnote{\arabic{footnote}}

\section{Introduction}

Figurative language is a challenge for both linguistic analysis \cite{dancygier2014figurative} and many natural language processing (NLP) tasks \cite{chakrabarty-etal-2022-rocket}. One representative task is literary translation \cite{karpinska-iyyer-2023-large}, where the translation of figurative language is one major difficulty. Among figurative language constructs, idioms are especially hard for a machine translation (MT) model due to their non-compositionality. For example, the meaning of the idiom ``bite the bullet'', deciding to do something difficult, is not simply composed of the meanings of ``bite'' and ``bullet''.

Of idioms in all languages, East Asian idioms constitute an interesting subset. Each of these idioms condenses its figurative meaning into a small number of characters, dominantly 4 characters (Chinese: \textit{sizichengyu}, Japanese: \textit{yojijukugo}, Korean: \textit{sajaseong-eo}, all literally meaning ``4-character idioms''). As the set of commonly accepted East Asian idioms and their meanings do \emph{not} change largely over time, the challenge of translating such idioms can seemingly be tackled by using a fixed list of literal and figurative candidate translations. This strategy has been commonly adopted by human translators \cite{tang2022petci} and MT researchers \cite{li2024translate} alike.

\begin{table}[t]
    \centering
    \small
    \begin{tabular}{l}
        \toprule
        \textbf{1. Idiom} \\
        \chinese{\textcolor{WildStrawberry}{刮目相看}}\\
        \midrule
        \textbf{2. Sentences} \\
        \chinese{小明的成绩提高得非常快，让老师和同学们}\\
        \chinese{都}\chinese{\textcolor{WildStrawberry}{刮目相看}}\chinese{。}\\
        \midrule
        \textbf{3. Context-Aware Translations} \\
        Xiaoming's grades soared impressively, \textcolor{WildStrawberry}{leaving} \\
        \textcolor{WildStrawberry}{both teachers and classmates in awe}. \\
        \midrule
        \textbf{Extracted Spans} (From Multiple Translations)\\
        \textbullet\ leaving both teachers and classmates in awe \\
        \textbullet\ taken everyone by surprise \\
        \textbullet\ like a phoenix reborn from its ashes \\
        \textbullet\ a blazing comet \\
        \textbullet\ earned everyone's admiration \\
        \textbullet\ with newfound respect \\
        \textbullet\ \ldots \\
        \midrule
        \textbf{Human Reference} (Sentences Not Available) \\
        \textbullet\ treat somebody with increased respect \\
        \textbullet\ look at somebody with new eyes \\
        \textbullet\ have a completely new appraisal of somebody \\
        \textbullet\ regard somebody with special esteem \\
        \bottomrule
    \end{tabular}
    \caption{\textbf{With our methods, we generate far more context-aware translations than human reference.} The pipeline of our context-aware translation is shown in Steps 1-3. To show results more clearly, we automatically extract the span (continuous words) in the translation that corresponds to the original idiom (Section \ref{sec:extension-to-a-larger-set}). More examples are available in Appendix \ref{sec:more-results}.}

    \label{tab:pipeline}
\end{table}

However, this approach has a major limitation. The existing East Asian idiom translation datasets provide translations out of context, i.e., idioms were translated without being incorporated into a surrounding sentence or paragraph. Hence, the translations sometimes require significant rewording to be appropriate in a given context. An example of such limitation is shown in Table \ref{tab:pipeline}. Despite that all 4 human reference translations are correct, the first 2 are awkward, as they overexaggerate a teacher's attitude towards a student as ``increased respect'' or ``special esteem'', and the last 2 use the active voice that interrupts the flow of the sentence.

In this work, we alleviate this limitation by using a SoTA large language model (LLM), GPT-4, to generate a dataset of context-aware idiom translations. We prompt GPT-4 to use different strategies to translate each idiom within various contexts. Moreover, to avoid accumulating translations by a brute-force and costly repetition of prompting, we select a small subset of Pareto-optimal prompting strategies from a comprehensive set, including zero-shot instructions inspired by human expertise and few-shot prompts that reuse high-quality translations. Table \ref{tab:pipeline} shows the steps that lead to a successful example, where our translations are superior in diversity and quality to the human reference. Our methods also beat commercial translation engines from Google and DeepL (Section \ref{sec:qualitative}).

\section{Method}
\label{sec:method}

In this section, we elaborate on the methods and experiment details for each step shown in Table \ref{tab:pipeline}.

\noindent \textbf{Step 1: Idioms} We obtain idioms from a dictionary for Chinese \cite{tang2022petci} and online resources for Japanese\footnote{\url{https://dictionary.goo.ne.jp/idiom/}} and Korean\footnote{\url{https://github.com/LiF-Lee/idioms/}}. These sources cover commonly used idioms in the 3 East-Asian languages. To test if our method generalizes to uncommon or new words that have an idiom-like structure, we also curate a set of plausible Chinese idioms. Plausible idioms are GPT-4-generated words which are not real idioms, but can fool GPT-4 when we ask it if the word is an idiom (Appendix \ref{sec:dictionaries}). For convenience, we use ``plausible Chinese'' to refer to the language of these idioms for convenience. For the 4 source languages, we limit the target language to English. To our best knowledge, the only human baseline that provides multiple translations for each idiom is the Chinese-English dictionary we use. 

\noindent \textbf{Step 2: Generate Sentences} \hspace{0.5em} To translate an idiom with context awareness, we need to translate a sentence that contains this idiom. Hence, we first generate multiple sentences containing a given idiom with GPT-4. For each of the 4 languages, we randomly sample 50 idioms and generate 10 sentences for each idiom, totalling 500 sentences.

\noindent \textbf{Step 3: Context-Aware Translation} \hspace{0.5em} Overall, we want multiple translations of each idiom within different contexts. This could be achieved if we only use a standard prompt (\textsc{Baseline}) to generate one translation per sentence. However, the contextual information is provided not only by the sentence but also by the paragraph that surrounds it. For example, a sentence can be translated more vividly when it appears in an everyday conversation than in a history book, but the \textsc{Baseline} translation is formal when no instructions are given (Table \ref{tab:naive}). To always have an option when a context is given, we generate multiple translations of each sentence by the following prompting strategies. Full prompts for each strategy can be found in Appendix \ref{sec:prompts}.

\begin{table}[htbp]
    \centering
    \small
    \begin{tabular}{l}
        \\  \toprule
        \textbf{Sentence} \\
        \chinese{他们通过}\chinese{\textcolor{WildStrawberry}{威逼利诱}}\chinese{，想要我放弃诉讼。}\\
         \midrule
        \textbf{{Baseline}} (History Book) \\
        They tried to get me to drop the lawsuit through \\
        \textcolor{WildStrawberry}{threats and inducements}.\\ \midrule
       \textbf{{Analogy Creative}} (Everyday Conversation) \\
        They tried to make me drop the lawsuit through \\
        \textcolor{WildStrawberry}{a carrot and stick approach}. \\
        \bottomrule
    \end{tabular}
    \caption{\textbf{The same idiom in different contexts (paragraphs) requires different translation strategies, even when the sentence is the same.} The two English sentences are translations of the same Chinese sentence. In all three sentences, the parts corresponding to the idiom is highlighted. Our pipeline is able to offer abundant choices for different contexts (in parentheses) by utilizing a comprehensive set of strategies (in bold).}
    \label{tab:naive}
\end{table}

Creativity is the key to adaptation in different contexts. To invoke creativity naively, we use a two-turn prompt that asks GPT-4 for 5 translations of one sentence (\textsc{Diversity Explicit}) and then for another 5 translations (\textsc{Diversity Dialog}). 

However, we should be able to get context-aware translations more efficiently. Instead of hoping for context-aware translations to be generated by sheer chance, we can directly ask for such translations. To do so, we explicitly ask GPT-4 to translate ``creatively'' (\textsc{Zero-Shot Creatively}).

Furthermore, we can add detailed instructions based on human expertise, instead of letting GPT-4 implicitly choose its translation strategy. Inspired by common translation strategies \cite{molina2002translation}, we use the following prompts: assuming a sentence appears in a paragraph of a certain genre (\textsc{Context Explicit}), using an analogy that is common (\textsc{Analogy Natural}) or uncommon (\textsc{Analogy Creative}), shuffling the order of clauses (\textsc{Shuffle Order}), rewriting the sentence without an idiom and then translating (\textsc{Two-Step}), avoiding using continuous spans (\textsc{Discontinuous 1}) or multi-word expressions (\textsc{Discontinuous 2}). For \textsc{Context Explicit}, we use 4 genres: a news report, a romance novel, an everyday conversation, and a history book.

While we can generate an abundance of translations, many are expected to be mundane or repetitive. Thus, to select a small subset that quickly helps human translators, we need to evaluate all translations to identify the best ones. Due to the low reproducibility and high cost of human evaluation, we instead prompt GPT-3.5 to score each sentence translation on a 1-5 scale based on faithfulness or creativity (Appendix \ref{sec:automatic-evaluation}). These two aspects are good proxies for the context-awareness of the idiom translation in the sentence. For each of the two aspects, the final score of a sentence translation is averaged from 5 runs, and the overall score of a prompting strategy is averaged from the scores of translations it produces. We use both aspects to select Pareto-optimal prompting strategies. 

After the initial round of evaluation, we reuse high-quality zero-shot translations as examples for few-shot prompting, prioritizing creativity. We choose the most creative translations and randomly sample from them to construct 5-shot prompts. The most creative translations are ones that score a 5 on creativity in at least 1 out of 5 runs. We use a 5-shot prompt with the word ``creatively'' (\textsc{Few-Shot Creatively}) or without (\textsc{Few-Shot}). The few-shot translations are evaluated using the same procedure for zero-shot translations.

Overall, we generate 27 translations per sentence with all prompting strategies, totalling 13,500 translations per language (Table \ref{tab:count-summary}). After identifying Pareto-optimal strategies, we further apply them to more idioms to expand our dataset (Section \ref{sec:extension-to-a-larger-set}).

For generation, we use the GPT-4 API and GPT-3.5 API from OpenAI\footnote{\texttt{gpt-4-0125-preview} and \texttt{gpt-3.5-turbo}}. Google\footnote{\texttt{translating-text-v3}} and DeepL are used as commercial translation engine baselines.  

In Appendix \ref{sec:resources}, we discuss alternative choices of resources, including dictionaries, parallel corpora, evaluation metrics, and LLMs. While we only experiment on the set of resources we choose above, we already reach the goal of benefiting translators by generating sufficiently high-quality translations.

\section{Results and Discussion}
\label{sec:results}

In this section, we first analyze the generated translations. Then, we examine how the optimal translation strategies work on a larger set of idioms. We also discuss how the dataset can be extended by an alternative pipeline using paragraph context.

\subsection{Quantitative Analysis}

To pick the Pareto-optimal translation strategies in each language, we aggregate the faithfulness and creativity scores of the translations from each strategy. The mean scores for Chinese idioms are visualized in Figure \ref{fig:chinese-scores}, while the visualizations for other languages and all numerical results are listed in Appendix \ref{sec:scores}. Here, we summarize 5 trends that generally hold true for all 4 languages. 

\begin{figure}
    \centering
    \includegraphics[width=\columnwidth]{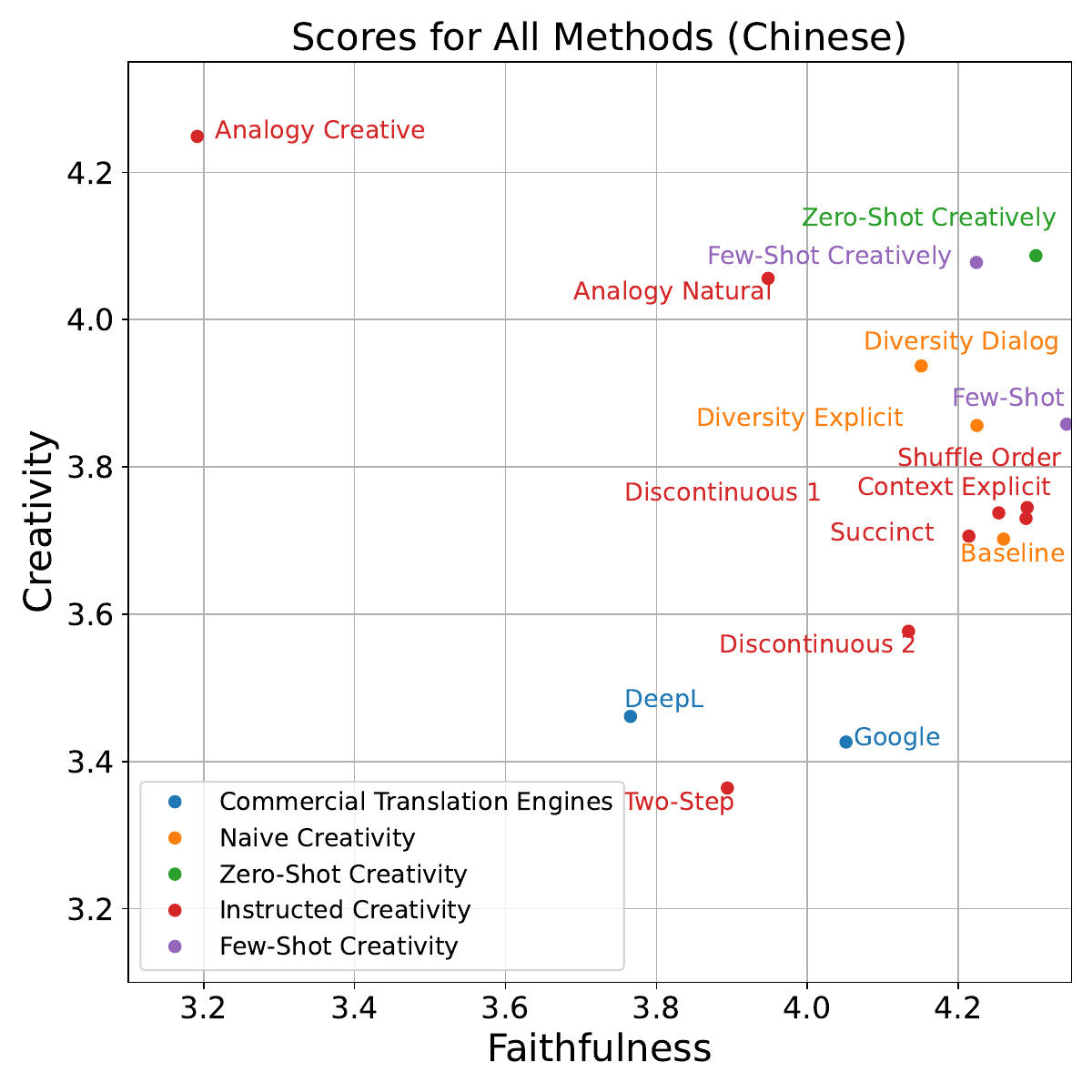}
    \caption{\textbf{Our strategies significantly differ in the mean faithfulness and creativity scores.} The strategies closer to the upper-right corner are better.}
    \label{fig:chinese-scores}
\end{figure}

First, GPT-4 is better at idiom translation than commercial translation engines. Both faithfulness and creativity are higher for GPT-4 translations than for Google and DeepL translations. This ranking aligns with our qualitative observations and supports the validity of our evaluation method.

Secondly, by naively invoking creativity of GPT-4 (\textsc{Diversity Explicit} and \textsc{Diversity Dialog}), we are able to improve creativity over the \textsc{Baseline}, at the cost of faithfulness. This result shows an inevitable trade-off between faithfulness and creativity without further instructions.

Thirdly, by simply adding the word ``creatively'' into the prompt (\textsc{Zero-Shot Creatively}), we are able to improve over the naive strategy and overcome the trade-off. This result motivates the search for stronger and more cost-efficient prompts, instead of repeatedly using weak prompts.

Fourthly, few strategies based on human expertise result in a Pareto improvement. This suggests that human expertise does not necessarily transfer to strong prompts, at least in the form of short prompts we use to briefly describe human translators' strategy. While a longer prompt with detailed instructions may bring out the full potential of a certain strategy, we do not consider these forms of prompts due to their high cost.

Finally, few-shot prompting strategies (\textsc{Few-Shot Creatively} and \textsc{Few-Shot}) are often Pareto-optimal. This result reveals the potential in using longer prompts and more sophisticated strategies to improve performance. However, from zero-shot to few-shot, the small improvement in scores costs many more tokens per idiom. 

\subsection{Qualitative Analysis}
\label{sec:qualitative}

We show translation examples in Table \ref{tab:pipeline} and Appendix \ref{sec:more-results}. For Chinese, we can compare our translations with the human reference. Thanks to the large number of different translations we get, most of them have not appeared in the dictionary. This shows that our method wins in diversity.

Regarding quality, GPT-4 almost always translate the idiom correctly\footnote{We look at more than a total of 10,000 translations by GPT-4 for over 100 random idioms, and we do not see obvious misinterpretation.}, and the translation quality are comparable to that of the human baseline (Table \ref{tab:pipeline}). In contrast, despite producing fluent sentences, Google and DeepL noticably misinterpret some idioms. For example, DeepL mistranslates the Chinese idiom ``\chinese{威逼利诱}'' (literally ``coercion and coaxing'') as ``bullying'' in the sentence in Table \ref{tab:naive}.

While all GPT-4-based strategies are able to produce faithful and creative translations, the proportions of such translations apparently differ for each strategy. In the GPT-4 translations, we observe two major failure patterns that cause a quality drop. First, GPT-4 fails to follow instructions on the translation strategy, producing idiom translations that are the same as the one from the \textsc{Baseline} prompt. While this behavior lowers the scores, these outputs are still valid translations, and outputs in undesired formats are rare (Appendix \ref{sec:cleaning-and-parsing}). Secondly, in the cases where instructions are fully followed, GPT-4 sometimes uses the strategy to improve the translation of other parts of the sentence, but not necessarily of the idiom itself. 

\subsection{Extension to a Larger Set}
\label{sec:extension-to-a-larger-set}

To validate our methods on a larger set of idioms, we apply 4 Pareto-optimal strategies (\textsc{Zero-Shot Creatively}, \textsc{Analogy Creative}, \textsc{Few-Shot}, and \textsc{Few-Shot Creatively}) on 500 top-frequency Chinese idioms (Appendix \ref{sec:parallel-corpora}). For the two few-shot methods, we reuse the most creative translations of the original 50 Chinese idioms. The numbers of Chinese idiom translations from all experiments are summarized in Table \ref{tab:count-summary}.

\begin{table}
  \centering
  \small
  \begin{tabular}{ll}
    \toprule
    \textbf{Item} & \textbf{Count} \\
    \midrule
    Idioms & $50$ \\
    Sentences & $50 \times 10 = 500$ \\
    Translations & $50 \times 10 \times 27 = 13500$ \\
    \midrule
    Idioms & $500$ \\
    Sentences & $500 \times 10 = 5000$ \\
    Translations & $500 \times 10 \times 4 = 20000$ \\
    \bottomrule
  \end{tabular}
  \caption{\textbf{The total number of idiom translations we generated for Chinese.} We first use all 27 translation strategies on 50 idioms. To expand the dataset, we then use 4 Pareto-optimal strategies on another 500 idioms.}
  \label{tab:count-summary}
\end{table}

Table \ref{tab:optimal-scores} shows that high scores are maintained for the new translations. We would like to further validate our evaluation strategy by comparing against the popular reference-free quality estimation (QE) metric \textsc{CometKiwi} \cite{rei-etal-2023-scaling}\footnote{\url{https://huggingface.co/Unbabel/wmt23-cometkiwi-da-xxl}}. Interestingly, the \textsc{CometKiwi} ranking is only consistent with the one given by our faithfulness score, suggesting the limitation of traditional QE metrics when creativity is among the evaluation criteria.

\begin{table}
  \centering
  \small
  \begin{tabular}{lccc}
    \toprule
    \textbf{Method} & \textbf{Faithfulness} & \textbf{Creativity} & \textsc{CometKiwi} \\
    \midrule
    \textsc{ZSC} & 4.27 $\pm$ 0.62 & 4.08 $\pm$ 0.36 & 0.79 $\pm$ 0.09 \\
    \textsc{AC} & 3.01 $\pm$ 1.02 & \textbf{4.27 $\pm$ 0.48} & 0.59 $\pm$ 0.14\\
    \textsc{FS} & \textbf{4.31 $\pm$ 0.63} & 3.80 $\pm$ 0.49 & \textbf{0.83 $\pm$ 0.08}\\
    \textsc{FSC} & 4.17 $\pm$ 0.71 & 4.07 $\pm$ 0.40 & 0.77 $\pm$ 0.12\\
    \bottomrule
  \end{tabular}
  \caption{\textbf{The scores (mean $\pm$ standard deviation) for the Pareto-optimal translation strategies (denoted by initials) are maintained on a larger set of idioms.} Faithfulness and creativity are in $[1, 5]$. \textsc{CometKiwi} is in $[0, 1]$. Highest (best) scores are boldfaced. The number of idiom translations from each method is 500 (sampled from 5,000). Faithfulness and \textsc{CometKiwi} give the same ranking. The strategy AC with the highest creativity is the lowest in faithfulness and \textsc{CometKiwi}.}
  \label{tab:optimal-scores}
\end{table}

\begin{figure}
    \centering
    \includegraphics[width=\columnwidth]{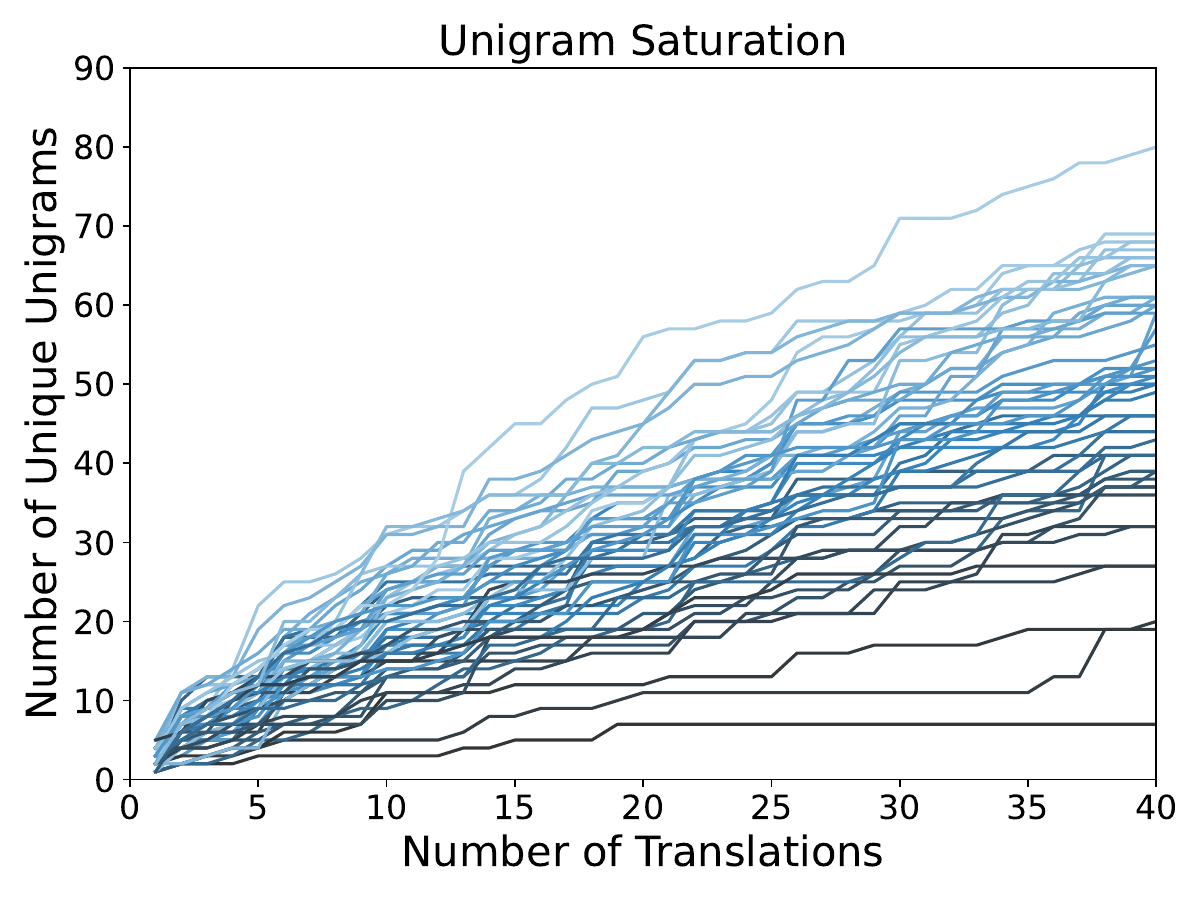}
    \caption{\textbf{The number of unique unigrams increases with the number of translations.} Increase rates and saturation points differ across 50 random idioms from 500 top-frequency Chinese idioms. Since there are 10 sentences for each idiom and 4 Pareto-optimal strategies, the total number of translations for each idiom is 40. For most idioms, the increase in the number of unique unigrams does not saturate at 40 translations.}
    \label{fig:unigram-saturation}
\end{figure}

Other than increasing the number of idioms, increasing the number of sentences containing each idiom is also a way to extend the dataset. Though we limit the number of sentences per idiom to 10, we observe that the number of different translations of each idiom keeps increasing with the number of sentences. To show this, we first extract the span in each translation that corresponds to the idiom (Appendix \ref{sec:spans}). Then, we use the number of unique unigrams in the spans as a proxy for the number of different translations for each idiom. We count unique unigrams instead of unique spans in order to avoid counting trivially different translations that only rearrange parts of other translations. For most idioms, the number of unique unigrams do not saturate as we increase the number of translations (Figure \ref{fig:unigram-saturation}). This illustrates the potential of scaling up our methods to more translations (sentences).

\subsection{Enhancing Context-Awareness}

Contrary to our expectations, we notice that the \textsc{Context Explicit} prompts often do not change the translation (one failure pattern in Section \ref{sec:qualitative}). Hence, for this specific strategy, we try to enhance context-awareness with a stronger pipeline. First, we specify different genres and generate multiple paragraphs that contain the sentence. Then, we ask GPT-4 to translate the paragraph, using the following 4 types of prompts. The first type is a baseline translation prompt. The second type encourages GPT-4 to emphasize 3 evaluation aspects, namely faithfulness, creativity, and word choices that match the theme. The third type utilizes step-by-step instructions that are generated by Auto-CoT \cite{liu-etal-2023-g}. The fourth type is a multi-turn dialog that asks GPT-4 to iteratively improve the translation based on each aspect. The full prompts can be found in Appendix \ref{sec:paragraph-translations-prompts}.

For each of the 4 languages, we choose 20 idioms and 1 sentence per idiom. We obtain a total of 800 translated paragraphs using 4 genres and 10 prompts. With the paragraph pipeline, we observe higher variation in translations from different genres than with our previous sentence pipeline. However, the different prompting strategies do not lead to meaningful variations for the same source paragraph (Appendix \ref{sec:paragraph-translations}). Due to the high cost of this pipeline, we leave the investigation of more idioms and prompting strategies as a future work.

\section{Related Work}

Different from traditional MT models, LLMs can produce diverse and less literal translations \cite{raunak-etal-2023-gpts}. Hence, a series of work has targeted generating diverse translations with LLMs and selecting the best. On one hand, translations can be generated from scratch using various prompting strategies. Then, ensembling methods can be applied to select the best candidates \cite{farinhas-etal-2023-empirical}. On the other hand, candidates can be further refined. Some examples include refining candidates generated by other machine translation systems \cite{raunak-etal-2023-leveraging}, iterative editing \cite{chen2024iterative, briakou2024translating}, self-correction \cite{feng-etal-2024-improving-language}, and multi-agent debate \cite{liang2023encouraging}. While existing work focused mostly on general translation, our work contributes in the more challenging task of generating diverse, high-quality translations for idioms.

\section{Conclusion}

In this work, we thoroughly test prompting strategies that generate different translations for an East Asian idiom. We identify the strategies that generate most creative and faithful translations. To our surprise, the prompts derived from human experience do not consistently generate quantitatively better translations. Finally, we use the Pareto optimal strategies to construct a dataset of high quality translations, which can help human translators.



\section*{Limitations}

Limitations of our work include:
\begin{itemize}
    \item \textbf{No external databases.} We discover that few-shot prompts could produce competitive translations, but are not able to use external translation as examples due to the lack of a high-quality and well-aligned parallel corpora.
    \item \textbf{No multi-agent or role-playing.} These orthogonal prompt-based directions provide effective methods that could possibly be combined with ours.
    \item \textbf{No language-specific strategies.} Our set of translation strategies is language-agnostic and thus not exhaustive.
    \item \textbf{No idiom categorization.} We do not consider the diverse linguistically motivated categorization of idioms in our general pipeline.
    \item \textbf{No expert evaluation.} We are not able to obtain the evaluation of translation quality from a large crowd of full-time professional translators due to cost and resource limits. Opinions from a smaller set of evaluators may carry personal biases, especially for the creative translation task we are investigating. Thus, we do not include human study in this work.
\end{itemize}
We would like to address these limitations in future work.

\bibliography{anthology, custom}

\appendix

\section{Resources}
\label{sec:resources}

In this section, we compare alternative resources with the ones we chose.

\subsection{Dictionaries}
\label{sec:dictionaries}

The dictionaries we used contain 4,310 idioms for Chinese, 2440 for Japanese, and 2,316 for Korean. These dictionaries are all available for public use. While other larger idiom dictionaries are available\footnote{\url{https://github.com/pwxcoo/chinese-xinhua/blob/master/data/idiom.json}}, our primary focus in this work was not a comprehensive coverage of all idioms in a language. The dictionaries we used should have included the idioms that are most frequently used.

It is worth noting that East Asian idioms include more than just the 4-character category. One example is the \textit{xiehouyu} in Chinese that has a different format. While we did not consider these other categories for experiments, our methods are applicable.

An interesting question is whether we can retrieve the idioms from GPT-4 just like we can retrieve the translations. To mine idioms, we asked GPT-4 to list a given number of idioms with a given initial. The initials are all valid \textit{pinyin} syllables in Chinese, and they can be written as one or more roman letters (e.g. ``a'', ``an'', and ``ang''). We obtained a list of 416 \textit{pinyin} syllables from Wikipedia\footnote{\url{https://en.wikipedia.org/wiki/Comparison\_of\_Standard\_Chinese\_transcription\_systems}}. One syllable might be a prefix of another, so different idiom lists returned by GPT-4 might partially overlap with each other. Based on various idiom dictionaries used in previous work in the NLP community \cite{li2024translate}, we estimated that the total number of frequently used Chinese idioms does not exceed 10K. Hence, for each initial, we asked GPT-4 to list 200 idioms. We expected the sufficiently large number of queries would give us a comprehensive list after deduplication. Furthermore, for each initial, we made 5 queries with different random seeds to improve stability, as we observed that GPT-4 produced results in some runs that were much worse than in other runs. 

We found that when listing idioms, GPT-4 tended to provide explanation or pronunciation of the idioms. Since these information are irrelevant for the listing task and significantly increase the output length, we asked GPT-4 to only list idioms without explaining them. We also found that when we did not explicitly require GPT-4 to list different idioms, GPT-4 tended to repeat a small set of idioms during listing. Hence, we explicitly asked GPT-4 to list different idioms.

Still, with the constraints in the prompt, GPT-4 occasionally produced undesirable content. We summarize the failure patterns below. 

First, not every query returned with 200 results. This can be expected, as there do not exist 200 idioms for some initials. However, we saw a large number of queries stopping at exactly 100 results. This indicated that GPT-4 was not interpreting the number in the instruction with full precision. 

Secondly, when given a certain initial, GPT-4 returned idioms with this initial in the beginning of the list, but idioms with different initials appeared later in the list. This was another example showing that the instruction was not precisely understood, even for the very simple task of listing. 

Thirdly, a majority of the returned expressions were not Chinese idioms. These fake idioms could be divided into two categories. For a idiom in the first category, when we asked GPT-4 whether this is a Chinese idiom, GPT-4 successfully identified the idiom to be fake. Similar to the previous failure pattern, this indicates that the classification ability of GPT-4 is weaker during listing than when classifying a single example. Some examples of the first category included general multi-word expressions (e.g. ``\chinese{半导体照明}'') and real idioms with a single character replaced (e.g. ``\chinese{按甲不动}'' from ``\chinese{按兵不动}''). The second category was more intriguing, as GPT-4 identified the fake idioms to be real (e.g. ``\chinese{落翅螳螂}''). In this category, the seemingly plausible idioms are constructed in a very similar way as real idioms. Hence, we manually selected 50 such idioms when we tested the translation strategies.

\subsection{Parallel Corpora}
\label{sec:parallel-corpora}

We have observed a low occurrence rate of idioms in large scale parallel corpora, for example the training set of WMT'23 \cite{kocmi-etal-2023-findings}. Idioms appear more often in literary text. However, due to copyright restrictions, the available literary parallel corpus is usually very small \cite{thai-etal-2022-exploring}. The largest literary parallel corpus we have found is the BWB corpus \cite{jiang-etal-2022-blonde}, which is a publicly available dataset of the English translation of Chinese web novels. Furthermore, due to sentence rearranging in literary translation, the smallest unit of a source-target pair in BWB is paragraphs. This makes it difficult to use BWB as a baseline to compare with the translations we get. Hence, we only used BWB to estimate the frequency ranking of idioms, where the frequency of an idiom is defined as the number of sentence pairs that contain this idiom. For the Chinese-English translation direction, GuoFeng \cite{wang-etal-2023-findings} is another large-scale literary parallel corpus. Since the dataset is also derived from web novels, we do not assume a large difference when the dataset is used for frequency estimation instead of BWB.

\subsection{Evaluation Metrics}

Translation quality estimation has long been studied in the NLP community. Currently, the popular metrics are \textsc{Comet} \cite{rei-etal-2020-comet}, \textsc{CometKiwi} \cite{rei-etal-2022-cometkiwi}, BLEURT \cite{sellam-etal-2020-bleurt}, BLEU \cite{papineni-etal-2002-bleu}, and chrF++ \cite{popovic-2017-chrf}\footnote{Both BLEU and chrF++ are implemented in SacreBLEU \cite{post-2018-call}.}. Among these metrics, the reference-free \textsc{CometKiwi} is the most suitable for a creative generation task. We used the \texttt{wmt-23-cometkiwi-da-xxl} version of \textsc{CometKiwi} \cite{rei-etal-2023-scaling}.

LLM-based metrics have been shown to perform well on text summarization and dialog generation \cite{liu-etal-2023-g}. For translation, LLMs were also applied to generate scores and textual evaluations \cite{kocmi-federmann-2023-gemba, fernandes-etal-2023-devil} based on MQM \cite{freitag-etal-2021-experts}. These work validated our choice of GPT-4 as an automatic evaluator of translation quality. We also explored the novel setting of evaluating creativity of translation, an aspect not covered by MQM.

Imperfect as they are, LLM-based automatic metrics are suitable for the assumed purpose of our dataset. We would like to provide a set of relatively good translations for a human translator to choose from. The automatic metrics reduce the cost of retrieving translations from LLMs and the time of human translators reading through the list. Automatic metrics may produce false negatives (high-quality translations that are excluded due to low scores) and false positives (low-quality translations that are chosen due to high scores). However, on the one hand, false negatives are not concerning due to the sheer number of translations we are able to retrieve. On the other hand, false positives can be easily identified by human translators. Hence, we chose automatic metrics to help data collection.

\subsection{Large Language Models}

LLMs other than GPT-4 have been widely used on translation-related tasks. Some examples are Claude-2\footnote{\url{https://www.anthropic.com/index/claude-2}}, Gemini-Pro\footnote{\url{https://cloud.google.com/vertex-ai/docs/generativeai/learn/models}}, Flan-T5 \cite{JMLR:v25:23-0870}, GPT-NeoX \cite{black-etal-2022-gpt}, and LLaMA \cite{touvron2023llama}. Since the prompting strategies we used in this work is model agnostic, it would be beneficial to use them on any model from the rapidly evolving set of LLMs. There is not a wide agreement on which model to choose. A consistent gap in translation quality between models was reported under some prompting strategies \cite{wang-li-2023-learning} and datasets but not others \cite{feng2024improving}.

\section{Prompts}
\label{sec:prompts}

In this section, we list all prompts we used. While a variety of prompts have been used for translation (Table \ref{tab:translation-prompt}), we used the prompts that more directly describes both our idiom translation task and relevant translation instructions. For all the prompts, we use a temperature of 1.0. In our pilot study, we observe that changing the temperature in the API call does not produce meaningful variations in the generated translations. The total cost of all GPT-related experiments in this paper, including pilot studies, was \$480.55.

\begin{table*}[htbp]
\centering
\small
\begin{tabular}{ll}
\toprule
\textbf{Reference} & \textbf{Strategies} \\
\midrule
\citet{jiao2023chatgpt} & ChatGPT generated templates \\
\citet{siu2023chatgpt} & Instructions in multi-turn dialogs \\
\citet{lyu-etal-2024-paradigm-shift} & Specifying a poetic style \\
\citet{he2023exploring} & Providing keywords, topics, or demonstrations mined by ChatGPT \\
\citet{na2024rethinking} & Using Skopos, functional equivalence, or text typology theory \\
\citet{tao2024catllm} & Specifying style by several word-level and sentence-level statistics \\
\citet{mu-etal-2023-augmenting} & Extracting most similar sentences from a translation database \\
\citet{yao2023empowering} & Providing the whole sentence and a literal translation of a cultural-specific entity \\
\bottomrule
\end{tabular}
\caption{\label{tab:translation-prompt}
A summary of prompting strategies for translation.
}
\end{table*}

\subsection{Idioms}

For Chinese idiom mining, we used the prompt:
\begin{itemize}
    \item Give 200 Chinese idioms that begin with \texttt{<PINYIN>}. Only list idioms. Do not explain them. No duplicates.
\end{itemize}
Here, \texttt{<PINYIN>} is chosen from a list of 416 \textit{pinyin} syllables obtained from Wikipedia\footnote{\url{https://en.wikipedia.org/wiki/Comparison\_of\_Standard\_Chinese\_transcription\_systems}}.

For checking if a result is a true Chinese idiom, we used the prompt:
\begin{itemize}
    \item Is \texttt{<IDIOM>} a Chinese idiom? Output yes or no.
\end{itemize}

For explanation generation, we used the prompt:
\begin{itemize}
    \item Is \texttt{<PLAUSIBLE IDIOM>} a Chinese idiom? Please explain.
\end{itemize}

\subsection{Sentences}
\label{sec:sentences}

For sentence generation, we used the prompt:
\begin{itemize}
    \item Can you make 10 \texttt{<LANGUAGE>} sentences with the \texttt{<LANGUAGE>} idiom \texttt{<IDIOM>}? Only list sentences. Do not explain.
\end{itemize}
For plausible Chinese, we use ``Chinese'' as the \texttt{<LANGUAGE>}.

\subsection{Zero-Shot Translations}

For zero-shot translation, we used the prompts in Table \ref{tab:zero-shot}.

\begin{table*}[htbp]
\small
\centering
\begin{tabular}{ll}
\toprule
\textbf{Name} & \textbf{Prompt} \\
\midrule
\textsc{Baseline} & \makecell[tl]{ Please translate the following sentence from \texttt{<LANGUAGE>} to English: \\ \texttt{<SENTENCE>}}\\
\midrule
\makecell[tl]{\textsc{Diversity} \\ \textsc{Explicit}} & \makecell[tl]{Please generate 5 different translations of the following sentence from \texttt{<LANGUAGE>} to English: \\ \texttt{<SENTENCE>}} \\
\midrule
\makecell[tl]{\textsc{Diversity} \\ \textsc{Dialog}} & \makecell[tl]{Please generate another 5 different translations.} \\
\midrule
\makecell[tl]{\textsc{Zero-Shot} \\ \textsc{Creatively}} & \makecell[tl]{Please creatively translate the following sentence from \texttt{<LANGUAGE>} to English:  \\ \texttt{<SENTENCE>}} \\
\midrule
\makecell[tl]{\textsc{Context} \\ \textsc{Explicit}} & \makecell[tl]{The sentence below comes from \texttt{<GENRE>}. Please translate it from \texttt{<LANGUAGE>} to English:\\ \texttt{<SENTENCE>}} \\
\midrule
\makecell[tl]{\textsc{Analogy} \\ \textsc{Natural}} & \makecell[tl]{ Please translate the following sentence from \texttt{<LANGUAGE>} to English: \\ \texttt{<SENTENCE>} \\
In the translation, please use an analogy commonly used in English.}\\
\midrule
\makecell[tl]{\textsc{Analogy} \\ \textsc{Creative}} & \makecell[tl]{ Please translate the following sentence from \texttt{<LANGUAGE>} to English: \\ \texttt{<SENTENCE>} \\
In the translation, please create a new analogy that has not been commonly used in English.}\\
\midrule
\makecell[tl]{\textsc{Shuffle} \\ \textsc{Order}} & \makecell[tl]{ Please translate the following sentence from \texttt{<LANGUAGE>} to English: \\ \texttt{<SENTENCE>} \\
Please try to change the order of clauses to make the translation more natural.}\\
\midrule
\makecell[tl]{\textsc{Succinct}} & \makecell[tl]{ Please translate the following sentence from \texttt{<LANGUAGE>} to English: \\ \texttt{<SENTENCE>} \\
Please translate the \texttt{<LANGUAGE>} idiom appeared in the sentence as succinctly as possible.}\\
\midrule
\makecell[tl]{\textsc{Two-Step}} & \makecell[tl]{ Please rewrite the following sentence in \texttt{<LANGUAGE>} without using a \texttt{<LANGUAGE>} idiom: \\ \texttt{<SENTENCE>} \\
Please translate the rewritten sentence to English.}\\
\midrule
\makecell[tl]{\textsc{Discontinuous 1}} & \makecell[tl]{ Please translate the following sentence from \texttt{<LANGUAGE>} to English: \\ \texttt{<SENTENCE>} \\
Please do not use a continuous span to translate the \texttt{<LANGUAGE>} idiom appeared in the sentence.}\\
\midrule
\makecell[tl]{\textsc{Discontinuous 2}} & \makecell[tl]{ Please translate the following sentence from \texttt{<LANGUAGE>} to English: \\ \texttt{<SENTENCE>} \\
Please do not use a multi-word expression to translate the \texttt{<LANGUAGE>} idiom appeared in the sentence.}\\
\bottomrule
\end{tabular}
\caption{\label{tab:zero-shot}
The prompts we used for zero-shot translation.
}
\end{table*}

\subsection{Few-Shot Translations}

For few-shot translation, we used the prompts in Table \ref{tab:few-shot}. The example sentence pairs are randomly chosen from a set of sentence pairs with highly creative translations.

\begin{table*}[htbp]
\small
\centering
\begin{tabular}{ll}
\toprule
\textbf{Name} & \textbf{Prompt} \\
\midrule
\textsc{Few-Shot} & \makecell[tl]{ Please translate the following sentences from \texttt{<LANGUAGE>} to English: \\ \texttt{<LANGUAGE>}: \texttt{<SOURCE 1>} \\ English: \texttt{<TARGET 1>} \\ \texttt{<LANGUAGE>}: \texttt{<SOURCE 2>} \\ English: \texttt{<TARGET 2>} \\ \texttt{<LANGUAGE>}: \texttt{<SOURCE 3>} \\ English: \texttt{<TARGET 3>} \\ \texttt{<LANGUAGE>}: \texttt{<SOURCE 4>} \\ English: \texttt{<TARGET 4>} \\ \texttt{<LANGUAGE>}: \texttt{<SOURCE 5>} \\ English: \texttt{<TARGET 5>} \\ \texttt{<LANGUAGE>}: \texttt{<SENTENCE>} \\ English:}\\
\midrule
\makecell[tl]{\textsc{Few-Shot} \\ \textsc{Creatively}} & \makecell[tl]{ Please creatively translate the following sentences from \texttt{<LANGUAGE>} to English: \\ \texttt{<LANGUAGE>}: \texttt{<SOURCE 1>} \\ English: \texttt{<TARGET 1>} \\ \texttt{<LANGUAGE>}: \texttt{<SOURCE 2>} \\ English: \texttt{<TARGET 2>} \\ \texttt{<LANGUAGE>}: \texttt{<SOURCE 3>} \\ English: \texttt{<TARGET 3>} \\ \texttt{<LANGUAGE>}: \texttt{<SOURCE 4>} \\ English: \texttt{<TARGET 4>} \\ \texttt{<LANGUAGE>}: \texttt{<SOURCE 5>} \\ English: \texttt{<TARGET 5>} \\ \texttt{<LANGUAGE>}: \texttt{<SENTENCE>} \\ English:}\\
\bottomrule
\end{tabular}
\caption{\label{tab:few-shot}
The prompts we used for few-shot translation.
}
\end{table*}

\subsection{Paragraph Translations}
\label{sec:paragraph-translations-prompts}

For instruction generation, we used the following 3 prompts:
\begin{itemize}
    \item If you are asked to translate a paragraph that contains a \texttt{<LANGUAGE>} idiom, what would you do to ensure that the translation of the idiom is faithful?
    \item If you are asked to translate a paragraph that contains a \texttt{<LANGUAGE>} idiom, what would you do to ensure that the translation of the idiom is creative?
    \item If you are asked to translate a paragraph that contains a \texttt{<LANGUAGE>} idiom, what would you do to ensure that the translation of the idiom matches the theme of its context?
\end{itemize}

For paragraph translation, we used the prompts in Table \ref{tab:paragraph}.

\begin{table*}[htbp]
\small
\centering
\begin{tabular}{ll}
\toprule
\textbf{Name} & \textbf{Prompt} \\
\midrule
\makecell[tl]{\textsc{Baseline}} & \makecell[tl]{Please translate the following paragraph from \texttt{<LANGUAGE>} to English. \\ \texttt{<PARAGRAPH>}}\\
\midrule
\makecell[tl]{\textsc{Faithful} \\ \textsc{Simple}} & \makecell[tl]{Please translate the following paragraph from \texttt{<LANGUAGE>} to English. \\ Please translate the idiom \texttt{<IDIOM>} faithfully. Do not explain. \\ \texttt{<PARAGRAPH>}}\\
\midrule
\makecell[tl]{\textsc{Creative} \\ \textsc{Simple}} & \makecell[tl]{Please translate the following paragraph from \texttt{<LANGUAGE>} to English. \\ Please translate the idiom \texttt{<IDIOM>} creatively. Do not explain. \\ \texttt{<PARAGRAPH>}}\\
\midrule
\makecell[tl]{\textsc{Theme} \\ \textsc{Simple}} & \makecell[tl]{Please translate the following paragraph from \texttt{<LANGUAGE>} to English. \\ Please translate the idiom \texttt{<IDIOM>} in a way that matches the theme. Do not explain. \\ \texttt{<PARAGRAPH>}}\\
\midrule
\makecell[tl]{\textsc{Faithful} \\ \textsc{CoT}} & \makecell[tl]{Please translate the following paragraph from \texttt{<LANGUAGE>} to English. \\ Please translate the idiom \texttt{<IDIOM>} faithfully. Do not explain. \\ \texttt{<PARAGRAPH>} \\ Please follow the instructions below: \\ \texttt{<FAITHFUL INSTRUCTIONS>}}\\
\midrule
\makecell[tl]{\textsc{Creative} \\ \textsc{CoT}} & \makecell[tl]{Please translate the following paragraph from \texttt{<LANGUAGE>} to English. \\ Please translate the idiom \texttt{<IDIOM>} creatively. Do not explain. \\ \texttt{<PARAGRAPH>} \\ Please follow the instructions below: \\ \texttt{<CREATIVE INSTRUCTIONS>}}\\
\midrule
\makecell[tl]{\textsc{Theme} \\ \textsc{CoT}} & \makecell[tl]{Please translate the following paragraph from \texttt{<LANGUAGE>} to English. \\ Please translate the idiom \texttt{<IDIOM>} in a way that matches the theme. Do not explain. \\ \texttt{<PARAGRAPH>} \\ Please follow the instructions below: \\ \texttt{<THEME INSTRUCTIONS>}}\\
\midrule
\makecell[tl]{\textsc{Faithful} \\ \textsc{Multi-Turn}} & \makecell[tl]{Please translate the following paragraph from \texttt{<LANGUAGE>} to English. \\ Please translate the idiom \texttt{<IDIOM>} faithfully. Do not explain. \\ \texttt{<PARAGRAPH>}}\\
\midrule
\makecell[tl]{\textsc{Creative} \\ \textsc{Multi-Turn}} & \makecell[tl]{Could you provide an alternative translation of the paragraph, where the idiom is translated more \\ creatively? The translation you provided has been widely used elsewhere.}\\
\midrule
\makecell[tl]{\textsc{Theme} \\ \textsc{Multi-Turn}} & \makecell[tl]{Could you provide an alternative translation of the paragraph, where the idiom is translated with words \\ that better match the context? The translation you provided can be used verbatim in a different context.}\\
\bottomrule
\end{tabular}
\caption{\label{tab:paragraph}
The prompts we used for paragraph translation.
}
\end{table*}

\subsection{Span Extraction}
\label{sec:span-extraction}

For span extraction, we used the following prompt:
\begin{itemize}
    \item Given the English translation of the \texttt{<LANGUAGE>} sentence, please only output the span that corresponds to the \texttt{<LANGUAGE>} idiom. \\
    \texttt{<LANGUAGE>} sentence: \texttt{<SOURCE>} \\
    English translation: \texttt{<TARGET>} \\
    \texttt{<LANGUAGE>} idiom: \texttt{<IDIOM>} \\
    Span: 
\end{itemize}
We only tested the prompt on Chinese sentences.

\subsection{Automatic Evaluation}
\label{sec:automatic-evaluation}

For automatic evaluation of faithfulness, we used the following prompt:
\begin{itemize}
    \item Please rate the faithfulness of the following idiom translation within a sentence.\\
    Idiom to be translated: \texttt{<IDIOM>} \\
    Original sentence containing this idiom: \texttt{<SOURCE>}\\
    Translation: \texttt{<TARGET>} \\
    Your faithfulness rating should be a score from 1 to 5, where 1 is not faithful at all and 5 is perfectly faithful. Return a single number as your rating.
\end{itemize}
For automatic evaluation of creativity, we used the following prompt:
\begin{itemize}
    \item Please rate the creativity of the following idiom translation within a sentence.\\
    Idiom to be translated: \texttt{<IDIOM>} \\
    Original sentence containing this idiom: \texttt{<SOURCE>}\\
    Translation: \texttt{<TARGET>} \\
    Your creativity rating should be a score from 1 to 5, where 1 is not creative at all (just plain language) and 5 is perfectly creative. Return a single number as your rating.
\end{itemize}

\section{Cleaning and Parsing}
\label{sec:cleaning-and-parsing}

In this section, we list the details for cleaning and parsing the model output.

\subsection{Sentences}

GPT-4 failed to generate sentences for very few idioms (4 out of all idioms). In these cases, GPT-4 was unable to identify the given idiom as a real word. Interestingly, this happened for real idioms, but not plausible idioms. For convenience of implementation, we save 10 empty sentences to the file when sentences are not generated.

Another failure pattern was that GPT-4 failed to include the idiom in the sentence. In these cases, GPT-4 split the idiom into two parts, or used synonyms to represent the meaning of the idiom. The statistics for all languages are listed in Table \ref{tab:cleaning-sentence}.

\begin{table*}
  \centering
  \small
  \begin{tabular}{lcccc}
    \toprule
    \textbf{Language} & \textbf{\# Idioms} & \textbf{\# Sentences} & \textbf{\# w/ Idiom} & \textbf{\% w/ Idiom} \\
    \midrule
    Chinese & 4310 & 43100 & 42873 & 99.71 \\
    Japanese & 2440 & 24400 & 24247 & 99.37 \\
    Korean & 2316 & 23160 & 22943 & 99.06 \\
    Plausible Chinese & 50 & 500 & 495 &99.00 \\
    \bottomrule
  \end{tabular}
  \caption{The statistics for noise in sentence generation.}
  \label{tab:cleaning-sentence}
\end{table*}

\subsection{Translations (Sentences)}

In general, the output translations are clean. In the cases where GPT-4 was prompted to generate multiple translations in a single response, we parsed the response to get the list of translations. For the succinct prompt, GPT-4 tended to provide an explanation, which we removed from the response to get the clean translation.

\subsection{Scores}

GPT-3.5 generated scores in different formats, including different prefixes and suffixes. We observed that all the irrelevant output can be cleaned by taking the first digit appearing in the response string as the score.

\subsection{Auto-CoT Instructions}

We repeatedly asked GPT-4 for translation instructions. Given the same prompt, GPT-4 returned different steps. The number of steps ranged from 6 to 8. The name and the description of each step also differed. However, each different set of steps from a single response was reasonable. Hence, for each combination of aspect and language, we only used one response as the Auto-CoT instruction.

\subsection{Paragraphs}

GPT-4 sometimes does not include the sentence verbatim in the paragraph. This is due to punctuation and language-specific phenomena, such as conjugation in Japanese and Korean. However, in most cases, the paragraph contains the idiom. The statistics for all languages are listed in Table \ref{tab:cleaning-paragraph}.

\begin{table*}
  \centering
  \small
  \begin{tabular}{lccccc}
    \toprule
    \textbf{Language} & \textbf{\# Paragraphs} & \textbf{\# w/ Idiom} & \textbf{\% w/ Idiom} & \textbf{\# w/ Sentence} & \textbf{\% w/ Sentence} \\
    \midrule
    Chinese & 80 & 80 & 100.00 & 51 & 63.75 \\
    Japanese & 80 & 79 & 98.75 & 37 & 46.25 \\ 
    Korean & 80 & 78 & 97.50 & 36 & 45.00 \\
    Plausible Chinese & 80 & 73 & 91.25 & 56 & 70.00 \\
    \bottomrule
  \end{tabular}
  \caption{The statistics for noise in paragraph generation.}
  \label{tab:cleaning-paragraph}
\end{table*}

\subsection{Translations (Paragraphs)}

No noise was observed.

\subsection{Spans}
\label{sec:spans}

GPT-4 was able to locate precisely the span in the translated sentence that corresponds to the idiom in the original sentence. The identified span is a substring of the translated sentence for 1994 out of 2000 Chinese-English sentence pairs, translated using the optimal strategies. The few failures were due to the change in capitalization and punctuation.

\section{More Results}
\label{sec:more-results}

In this section, we show more examples from our results. No spans are highlighted for these examples, as we did not perform manual labeling and did not run the span queries (Appendix \ref{sec:span-extraction}) for these translations. The total number of sentence translations is 13,500 for each language. For Chinese, we further generate 20,000 translations with Pareto-optimal strategies. For each translation, we generate 5 faithfulness and 5 creativity scores.

\subsection{Sentence Translations}
\label{sec:sentence-translations}

We show more translations for all languages we used in Tables \ref{tab:sentence-translations-chinese} (Chinese), \ref{tab:sentence-translations-japanese} (Japanese), \ref{tab:sentence-translations-korean} (Korean), and \ref{tab:sentence-translations-plausible-chinese} (plausible Chinese). All other translations can be found in the published data. The total number of paragraph translations is 800 for each language.

\begin{table*}
\centering
\tiny
\begin{tabular}{clcc}
\toprule
\textbf{Method} & \textbf{Sentence} & \textbf{Faithfulness} &\textbf{Creativity} \\
\midrule
\textsc{Original} & \chinese{他们通过威逼利诱，想要我放弃诉讼。} & N/A & N/A \\
\textsc{Google} & \multicolumn{1}{p{0.6\linewidth}}{\raggedright They used threats and inducements to try to get me to drop the lawsuit.} & 4.8 & 3.8 \\
\textsc{DeepL} & \multicolumn{1}{p{0.6\linewidth}}{\raggedright They wanted me to drop the lawsuit through bullying.} & 2.2 & 2.6 \\
\textsc{Baseline} & \multicolumn{1}{p{0.6\linewidth}}{\raggedright They tried to get me to drop the lawsuit through threats and inducements.} & 4.2 & 3.8 \\
\textsc{Diversity Explicit} & \multicolumn{1}{p{0.6\linewidth}}{\raggedright Through intimidation and temptation, they want me to give up the lawsuit.} & 5.0 & 3.6 \\
\textsc{Diversity Explicit} & \multicolumn{1}{p{0.6\linewidth}}{\raggedright They employed threats and bribes, trying to get me to abandon the lawsuit.} & 4.4 & 4.0 \\
\textsc{Diversity Explicit} & \multicolumn{1}{p{0.6\linewidth}}{\raggedright By wielding both threats and incentives, they aim to make me drop the lawsuit.} & 4.4 & 4.0 \\
\textsc{Diversity Explicit} & \multicolumn{1}{p{0.6\linewidth}}{\raggedright They attempted to make me forsake the lawsuit by means of coercion and enticement.} & 4.4 & 3.2 \\
\textsc{Diversity Explicit} & \multicolumn{1}{p{0.6\linewidth}}{\raggedright They're using a mix of scare tactics and promises in hopes I'll relinquish the lawsuit.} & 4.0 & 4.2 \\
\textsc{Diversity Dialog} & \multicolumn{1}{p{0.6\linewidth}}{\raggedright They are using a combination of threats and rewards to persuade me to withdraw from the lawsuit.} & 4.6 & 4.0 \\
\textsc{Diversity Dialog} & \multicolumn{1}{p{0.6\linewidth}}{\raggedright With a strategy of intimidation and allure, they seek to have me forego the lawsuit.} & 4.0 & 4.0 \\
\textsc{Diversity Dialog} & \multicolumn{1}{p{0.6\linewidth}}{\raggedright They're applying pressure and offering inducements, hoping I will abandon my legal action.} & 4.8 & 4.0 \\
\textsc{Diversity Dialog} & \multicolumn{1}{p{0.6\linewidth}}{\raggedright By employing coercion and temptation, their goal is for me to give up my lawsuit.} & 4.2 & 3.0 \\
\textsc{Diversity Dialog} & \multicolumn{1}{p{0.6\linewidth}}{\raggedright Through both menacing and luring tactics, they wish for me to relinquish the lawsuit.} & 4.4 & 4.0 \\
\textsc{Zero-Shot Creatively} & \multicolumn{1}{p{0.6\linewidth}}{\raggedright Through a mix of threats and temptations, they sought to make me abandon my lawsuit.} & 4.6 & 4.0 \\
\textsc{Context Explicit} & \multicolumn{1}{p{0.6\linewidth}}{\raggedright They tried to make me give up the lawsuit through threats and bribes.} & 4.4 & 2.8 \\
\textsc{Context Explicit} & \multicolumn{1}{p{0.6\linewidth}}{\raggedright They, through intimidation and inducement, want me to give up the lawsuit.} & 4.6 & 3.0 \\
\textsc{Context Explicit} & \multicolumn{1}{p{0.6\linewidth}}{\raggedright They tried to make me give up the lawsuit through threats and bribes.} & 4.6 & 3.0 \\
\textsc{Context Explicit} & \multicolumn{1}{p{0.6\linewidth}}{\raggedright They tried to make me give up the lawsuit through intimidation and bribery.} & 4.6 & 3.2 \\
\textsc{Analogy Natural} & \multicolumn{1}{p{0.6\linewidth}}{\raggedright They tried to make me drop the lawsuit through a carrot and stick approach.} & 4.0 & 4.0 \\
\textsc{Analogy Creative} & \multicolumn{1}{p{0.6\linewidth}}{\raggedright They tried to make me drop the lawsuit by dangling carrots laced with threats, like offering honey wrapped in thorns.} & 3.6 & 4.2 \\
\textsc{Shuffle Order} & \multicolumn{1}{p{0.6\linewidth}}{\raggedright They tried to get me to drop the lawsuit by using threats and bribes.} & 4.0 & 3.8 \\
\textsc{Succinct} & \multicolumn{1}{p{0.6\linewidth}}{\raggedright They, through threats and bribes, want me to give up the lawsuit.} & 4.2 & 3.0 \\
\textsc{Two-Step} & \multicolumn{1}{p{0.6\linewidth}}{\raggedright They used threats and temptations, trying to make me give up the lawsuit.} & 4.0 & 3.6 \\
\textsc{Discontinuous 1} & \multicolumn{1}{p{0.6\linewidth}}{\raggedright They, through threats and promises, want me to give up the lawsuit.} & 4.8 & 3.0 \\
\textsc{Discontinuous 2} & \multicolumn{1}{p{0.6\linewidth}}{\raggedright They want me to give up the lawsuit by coercing and bribing.} & 4.0 & 2.8 \\
\textsc{Few-Shot} & \multicolumn{1}{p{0.6\linewidth}}{\raggedright They tried to make me give up the lawsuit through threats and bribes.} & 4.6 & 3.2 \\
\textsc{Few-Shot Creatively} & \multicolumn{1}{p{0.6\linewidth}}{\raggedright They attempted to sway my resolve through a mix of threats and temptations, hoping I would abandon the lawsuit.} & 5.0 & 4.0 \\
\bottomrule
\end{tabular}
\caption{The example translations for Chinese.}
\label{tab:sentence-translations-chinese}
\end{table*}

\begin{table*}
\centering
\tiny
\begin{tabular}{clcc}
\toprule
\textbf{Method} & \textbf{Sentence} & \textbf{Faithfulness} &\textbf{Creativity} \\
\midrule
\textsc{Original} & \japanese{私たちの小さなプロジェクトが、まさに燎原之火となり、急速に広がりました} & N/A & N/A \\
\textsc{Google} & \multicolumn{1}{p{0.6\linewidth}}{\raggedright Our small project became a wildfire and spread rapidly.} & 4.2 & 4.0 \\
\textsc{DeepL} & \multicolumn{1}{p{0.6\linewidth}}{\raggedright Our little project became a veritable bonfire and spread rapidly.} & 2.8 & 4.0 \\
\textsc{Baseline} & \multicolumn{1}{p{0.6\linewidth}}{\raggedright Our small project, just like a spark in a field, rapidly spread far and wide.} & 3.4 & 4.2 \\
\textsc{Diversity Explicit} & \multicolumn{1}{p{0.6\linewidth}}{\raggedright Our small project has indeed become a spark that quickly spread like wildfire.} & 4.4 & 4.2 \\
\textsc{Diversity Explicit} & \multicolumn{1}{p{0.6\linewidth}}{\raggedright Our modest project has truly become a blaze that rapidly expanded.} & 3.6 & 3.8 \\
\textsc{Diversity Explicit} & \multicolumn{1}{p{0.6\linewidth}}{\raggedright The little project of ours has indeed turned into a raging fire, spreading swiftly.} & 4.0 & 4.0 \\
\textsc{Diversity Explicit} & \multicolumn{1}{p{0.6\linewidth}}{\raggedright Our small project has indeed become a prairie fire, spreading rapidly.} & 4.8 & 4.2 \\
\textsc{Diversity Explicit} & \multicolumn{1}{p{0.6\linewidth}}{\raggedright Our minor project quickly became a spark that turned into a widespread inferno.} & 2.8 & 4.0 \\
\textsc{Diversity Dialog} & \multicolumn{1}{p{0.6\linewidth}}{\raggedright Our little project has truly blossomed into a widespread blaze, spreading quickly.} & 2.8 & 3.8 \\
\textsc{Diversity Dialog} & \multicolumn{1}{p{0.6\linewidth}}{\raggedright Our small initiative indeed became a conflagration that rapidly expanded.} & 4.4 & 4.0 \\
\textsc{Diversity Dialog} & \multicolumn{1}{p{0.6\linewidth}}{\raggedright The small project we started has really caught on like wildfire, spreading rapidly.} & 4.4 & 4.0 \\
\textsc{Diversity Dialog} & \multicolumn{1}{p{0.6\linewidth}}{\raggedright Our humble project turned into a blazing fire, quickly covering ground.} & 4.4 & 4.0 \\
\textsc{Diversity Dialog} & \multicolumn{1}{p{0.6\linewidth}}{\raggedright Our tiny project has, indeed, turned into a spark that has rapidly spread far and wide.} & 3.6 & 4.0 \\
\textsc{Zero-Shot Creatively} & \multicolumn{1}{p{0.6\linewidth}}{\raggedright Our humble project, akin to a spark in a dry field, swiftly ignited, spreading far and wide with rapid fervor.} & 4.4 & 4.2 \\
\textsc{Context Explicit} & \multicolumn{1}{p{0.6\linewidth}}{\raggedright Our small project has indeed become a spark that quickly spread like wildfire.} & 4.2 & 4.2 \\
\textsc{Context Explicit} & \multicolumn{1}{p{0.6\linewidth}}{\raggedright Our small project became like a spark in a prairie, spreading rapidly.} & 3.2 & 4.0 \\
\textsc{Context Explicit} & \multicolumn{1}{p{0.6\linewidth}}{\raggedright Our small project has indeed become a spark that ignites the plains, spreading rapidly.} & 3.2 & 4.2 \\
\textsc{Context Explicit} & \multicolumn{1}{p{0.6\linewidth}}{\raggedright Our small project became like a spark in a prairie, spreading rapidly.} & 3.4 & 4.0 \\
\textsc{Analogy Natural} & \multicolumn{1}{p{0.6\linewidth}}{\raggedright Our little project caught fire like wildfire and spread rapidly.} & 3.6 & 4.2 \\
\textsc{Analogy Creative} & \multicolumn{1}{p{0.6\linewidth}}{\raggedright Our modest project, just like a spark in a sea of whispers, quickly spread far and wide.} & 1.6 & 4.2 \\
\textsc{Shuffle Order} & \multicolumn{1}{p{0.6\linewidth}}{\raggedright Our small project has indeed become a blaze that quickly spread far and wide.} & 4.0 & 3.6 \\
\textsc{Succinct} & \multicolumn{1}{p{0.6\linewidth}}{\raggedright Our small project has indeed become a spark that started a prairie fire, spreading rapidly.} & 5.0 & 4.4 \\
\textsc{Two-Step} & \multicolumn{1}{p{0.6\linewidth}}{\raggedright Our small project really spread rapidly.} & 1.8 & 2.0 \\
\textsc{Discontinuous 1} & \multicolumn{1}{p{0.6\linewidth}}{\raggedright Our small project, just like a spark in a field, rapidly spread far and wide.} & 3.4 & 4.2 \\
\textsc{Discontinuous 2} & \multicolumn{1}{p{0.6\linewidth}}{\raggedright Our small project has indeed become a wildfire and spread rapidly.} & 4.2 & 3.8 \\
\textsc{Few-Shot} & \multicolumn{1}{p{0.6\linewidth}}{\raggedright Our small project became like a wildfire, spreading rapidly.} & 4.6 & 4.0 \\
\textsc{Few-Shot Creatively} & \multicolumn{1}{p{0.6\linewidth}}{\raggedright Our modest project became a wildfire, spreading rapidly far and wide.} & 4.0 & 4.0 \\
\bottomrule
\end{tabular}
\caption{The example translations for Japanese.}
\label{tab:sentence-translations-japanese}
\end{table*}

\begin{table*}
\centering
\tiny
\begin{tabular}{clcc}
\toprule
\textbf{Method} & \textbf{Sentence} & \textbf{Faithfulness} &\textbf{Creativity} \\
\midrule
\textsc{Original} & \korean{진짜 그 사람이 싫어서 안 만나는 건지, 아니면 담인인지 모르겠어.} & N/A & N/A \\
\textsc{Google} & \multicolumn{1}{p{0.6\linewidth}}{\raggedright I don't know if I'm not dating that person because I really hate him, or if he's just a temperamental person.} & 2.2 & 3.6 \\
\textsc{DeepL} & \multicolumn{1}{p{0.6\linewidth}}{\raggedright I don't know if I'm not seeing him because I really don't like him, or if he's a dummy.} & 2.0 & 3.8 \\
\textsc{Baseline} & \multicolumn{1}{p{0.6\linewidth}}{\raggedright I'm not sure if it's really because I dislike that person that I don't meet them, or if it's because I'm just busy.} & 3.2 & 3.0 \\
\textsc{Diversity Explicit} & \multicolumn{1}{p{0.6\linewidth}}{\raggedright I'm not sure if it's because I really dislike that person or if I'm just being indifferent.} & 3.6 & 3.8 \\
\textsc{Diversity Explicit} & \multicolumn{1}{p{0.6\linewidth}}{\raggedright I can't tell if I'm not meeting them because I really hate them, or if I'm just unbothered.} & 3.6 & 4.0 \\
\textsc{Diversity Explicit} & \multicolumn{1}{p{0.6\linewidth}}{\raggedright It's unclear whether I'm avoiding them because I genuinely don't like them, or if I'm just being nonchalant.} & 3.0 & 4.0 \\
\textsc{Diversity Explicit} & \multicolumn{1}{p{0.6\linewidth}}{\raggedright I'm uncertain if my not seeing them is due to real distaste or simply indifference.} & 4.0 & 4.0 \\
\textsc{Diversity Explicit} & \multicolumn{1}{p{0.6\linewidth}}{\raggedright It's hard to say whether my avoidance is because of genuine dislike or if I am just indifferent.} & 4.0 & 4.0 \\
\textsc{Diversity Dialog} & \multicolumn{1}{p{0.6\linewidth}}{\raggedright I can't figure out if my refusal to meet is because I truly detest that person, or if it's just apathy.} & 4.2 & 3.6 \\
\textsc{Diversity Dialog} & \multicolumn{1}{p{0.6\linewidth}}{\raggedright I'm unsure whether my avoidance is rooted in real animosity towards that person or mere indifference.} & 4.4 & 4.0 \\
\textsc{Diversity Dialog} & \multicolumn{1}{p{0.6\linewidth}}{\raggedright It's difficult to determine if I'm not seeing them because I genuinely loathe them, or if I'm just detached.} & 3.6 & 4.0 \\
\textsc{Diversity Dialog} & \multicolumn{1}{p{0.6\linewidth}}{\raggedright I'm puzzled whether my lack of desire to meet stems from actual aversion or simply from being disinterested.} & 3.8 & 4.2 \\
\textsc{Diversity Dialog} & \multicolumn{1}{p{0.6\linewidth}}{\raggedright I find it challenging to discern if my avoidance is due to a genuine dislike or if it's just a matter of being indifferent.} & 4.2 & 4.0 \\
\textsc{Zero-Shot Creatively} & \multicolumn{1}{p{0.6\linewidth}}{\raggedright I can't tell if they're genuinely avoiding that person because they dislike them, or if it's because they're shy.} & 3.8 & 4.0 \\
\textsc{Context Explicit} & \multicolumn{1}{p{0.6\linewidth}}{\raggedright I'm not sure if it's because they really dislike that person that they're not meeting, or if it's because they're just playing hard to get.} & 4.0 & 4.2 \\
\textsc{Context Explicit} & \multicolumn{1}{p{0.6\linewidth}}{\raggedright I can't tell if it's because they really dislike that person that they're not meeting, or if it's because they're playing hard to get.} & 3.8 & 3.8 \\
\textsc{Context Explicit} & \multicolumn{1}{p{0.6\linewidth}}{\raggedright I can't tell if they're really avoiding meeting because they dislike that person, or if they're just busy.} & 3.6 & 3.4 \\
\textsc{Context Explicit} & \multicolumn{1}{p{0.6\linewidth}}{\raggedright I'm not sure if it's because they really dislike that person or if it's because they are just busy.} & 3.6 & 3.6 \\
\textsc{Analogy Natural} & \multicolumn{1}{p{0.6\linewidth}}{\raggedright "I can't tell if they're really avoiding someone because they dislike them, or if they're just playing hard to get."} & 3.4 & 4.0 \\
\textsc{Analogy Creative} & \multicolumn{1}{p{0.6\linewidth}}{\raggedright "I can't tell if they're really avoiding meeting because they dislike that person, or if it's like trying to decipher the flavor of a mystery soda."} & 2.2 & 4.4 \\
\textsc{Shuffle Order} & \multicolumn{1}{p{0.6\linewidth}}{\raggedright I can't tell if they're really avoiding meeting because they dislike that person, or if they're just busy.} & 3.8 & 3.4 \\
\textsc{Succinct} & \multicolumn{1}{p{0.6\linewidth}}{\raggedright I can't tell if it's really because they dislike that person that they're not meeting up, or if they're just playing hard to get. } & 4.0 & 4.0 \\
\textsc{Two-Step} & \multicolumn{1}{p{0.6\linewidth}}{\raggedright I don't know if it's really because I dislike that person that I'm not meeting them, or if there is another reason.} & 4.2 & 3.2 \\
\textsc{Discontinuous 1} & \multicolumn{1}{p{0.6\linewidth}}{\raggedright I can't tell if they're really not meeting because they dislike that person, or if it's because they're playing hard to get.} & 3.8 & 4.0 \\
\textsc{Discontinuous 2} & \multicolumn{1}{p{0.6\linewidth}}{\raggedright I'm not sure if it's really because I don't like that person that I'm not meeting them, or if it's something else.} & 4.2 & 3.6 \\
\textsc{Few-Shot} & \multicolumn{1}{p{0.6\linewidth}}{\raggedright I can't tell if they're not meeting because they genuinely dislike the person or if it's just an excuse.} & 4.0 & 4.0 \\
\textsc{Few-Shot Creatively} & \multicolumn{1}{p{0.6\linewidth}}{\raggedright I can't tell if it's genuinely because they dislike that person that they don't meet up, or if it's just an excuse.} & 4.0 & 3.8 \\
\bottomrule
\end{tabular}
\caption{The example translations for Korean.}
\label{tab:sentence-translations-korean}
\end{table*}

\begin{table*}
\centering
\tiny
\begin{tabular}{clcc}
\toprule
\textbf{Method} & \textbf{Sentence} & \textbf{Faithfulness} &\textbf{Creativity} \\
\midrule
\textsc{Original} & \chinese{每年的春天，园子里春荷盛开，秋天则是秋菊竞艳。} & N/A & N/A \\
\textsc{Google} & \multicolumn{1}{p{0.6\linewidth}}{\raggedright Every spring, lotus flowers bloom in the garden, and in autumn, chrysanthemums bloom.} & 3.2 & 2.2 \\
\textsc{DeepL} & \multicolumn{1}{p{0.6\linewidth}}{\raggedright Every spring, the garden is in full bloom with spring lotus, and in the fall, fall chrysanthemums compete.} & 4.2 & 4.0 \\
\textsc{Baseline} & \multicolumn{1}{p{0.6\linewidth}}{\raggedright Every spring, the garden is filled with blooming spring lotuses, while in autumn, it's the chrysanthemums that vie for beauty.} & 4.0 & 4.0 \\
\textsc{Diversity Explicit} & \multicolumn{1}{p{0.6\linewidth}}{\raggedright Every spring, the garden blooms with spring lotuses, while in autumn, the chrysanthemums vie for beauty.} & 4.2 & 3.6 \\
\textsc{Diversity Explicit} & \multicolumn{1}{p{0.6\linewidth}}{\raggedright In the garden, spring sees the lush blossoming of lotuses, and autumn showcases the competitive bloom of chrysanthemums.} & 4.0 & 4.0 \\
\textsc{Diversity Explicit} & \multicolumn{1}{p{0.6\linewidth}}{\raggedright Yearly, spring brings the luxuriant opening of lotuses in the garden, and autumn displays the dazzling chrysanthemums.} & 3.2 & 3.8 \\
\textsc{Diversity Explicit} & \multicolumn{1}{p{0.6\linewidth}}{\raggedright Every year, the springtime garden is filled with blooming lotuses, and autumn with the splendid display of chrysanthemums.} & 3.6 & 3.8 \\
\textsc{Diversity Explicit} & \multicolumn{1}{p{0.6\linewidth}}{\raggedright The garden annually celebrates spring with the flourishing of lotuses, and autumn with the striking rivalry of chrysanthemums.} & 3.8 & 4.0 \\
\textsc{Diversity Dialog} & \multicolumn{1}{p{0.6\linewidth}}{\raggedright Each spring, the garden is adorned with burgeoning lotuses, while autumn brings forth the competing beauty of chrysanthemums.} & 4.4 & 4.0 \\
\textsc{Diversity Dialog} & \multicolumn{1}{p{0.6\linewidth}}{\raggedright Annually, the spring season witnesses the garden's lotuses in full bloom, followed by the chrysanthemums' vibrant display in autumn.} & 4.0 & 4.2 \\
\textsc{Diversity Dialog} & \multicolumn{1}{p{0.6\linewidth}}{\raggedright The garden experiences a bounty of spring lotuses every year, and in autumn, the chrysanthemums burst into competitive radiance.} & 3.6 & 3.8 \\
\textsc{Diversity Dialog} & \multicolumn{1}{p{0.6\linewidth}}{\raggedright Every spring, lotuses thrive in the garden, with autumn presenting a spectacle of chrysanthemums competing in splendor.} & 3.8 & 4.0 \\
\textsc{Diversity Dialog} & \multicolumn{1}{p{0.6\linewidth}}{\raggedright In the garden, each year, spring is marked by the blossoming of lotuses and autumn by the riotous beauty of chrysanthemums in competition.} & 4.4 & 4.0 \\
\textsc{Zero-Shot Creatively} & \multicolumn{1}{p{0.6\linewidth}}{\raggedright Every spring, the garden blooms with vibrant spring lotus, while in autumn, the chrysanthemums vie in beauty.} & 3.8 & 4.0 \\
\textsc{Context Explicit} & \multicolumn{1}{p{0.6\linewidth}}{\raggedright Every spring, the lotus blooms abundantly in the garden, while in autumn, the chrysanthemums vie in beauty.} & 4.0 & 4.0 \\
\textsc{Context Explicit} & \multicolumn{1}{p{0.6\linewidth}}{\raggedright Every spring, the lotus blooms abundantly in the garden, while in autumn, the chrysanthemums compete in beauty.} & 4.0 & 4.0 \\
\textsc{Context Explicit} & \multicolumn{1}{p{0.6\linewidth}}{\raggedright Every spring, the lotus blooms abundantly in the garden, while in autumn, the chrysanthemums compete in beauty.} & 4.0 & 4.0 \\
\textsc{Context Explicit} & \multicolumn{1}{p{0.6\linewidth}}{\raggedright Every spring, the lotus blooms profusely in the garden, while in autumn, the chrysanthemums vie in beauty.} & 3.6 & 3.8 \\
\textsc{Analogy Natural} & \multicolumn{1}{p{0.6\linewidth}}{\raggedright Every spring, the garden blooms with spring lotuses, much like a sea of blossoms, and in autumn, autumn chrysanthemums vie in beauty, akin to a painter's vibrant palette.} & 3.8 & 4.2 \\
\textsc{Analogy Creative} & \multicolumn{1}{p{0.6\linewidth}}{\raggedright In the springtime each year, the garden blossoms into a sea of spring lotuses, while in autumn, it transforms into a stage where autumn chrysanthemums vie for the spotlight.} & 3.6 & 4.2 \\
\textsc{Shuffle Order} & \multicolumn{1}{p{0.6\linewidth}}{\raggedright Every spring, the lotus blossoms flourish in the garden, while in autumn, the chrysanthemums vie in beauty.} & 4.0 & 4.0 \\
\textsc{Succinct} & \multicolumn{1}{p{0.6\linewidth}}{\raggedright Every spring, the lotus flowers bloom in abundance in the garden, and in autumn, the chrysanthemums vie in beauty.} & 3.6 & 3.8 \\
\textsc{Two-Step} & \multicolumn{1}{p{0.6\linewidth}}{\raggedright Every spring, the lotuses in the garden bloom in abundance; come autumn, it is the chrysanthemums' turn to bloom competitively.} & 3.6 & 4.0 \\
\textsc{Discontinuous 1} & \multicolumn{1}{p{0.6\linewidth}}{\raggedright Every spring, the garden is in full bloom with spring lotuses, and in autumn, autumn chrysanthemums vie in beauty.} & 3.6 & 3.8 \\
\textsc{Discontinuous 2} & \multicolumn{1}{p{0.6\linewidth}}{\raggedright Every spring, the garden is full of blooming lotuses, and in autumn, autumn chrysanthemums vie in beauty.} & 3.6 & 4.0 \\
\textsc{Few-Shot} & \multicolumn{1}{p{0.6\linewidth}}{\raggedright Every spring, the garden blossoms with spring lotuses, while in autumn, the autumn chrysanthemums vie in beauty.} & 4.0 & 4.0 \\
\textsc{Few-Shot Creatively} & \multicolumn{1}{p{0.6\linewidth}}{\raggedright Every spring, the garden is a spectacle of blooming spring lotuses, while autumn boasts a competitive display of vibrant chrysanthemums.} & 4.0 & 4.2 \\
\bottomrule
\end{tabular}
\caption{The example translations for plausible Chinese.}
\label{tab:sentence-translations-plausible-chinese}
\end{table*}

\subsection{Paragraph Translations}
\label{sec:paragraph-translations}

For the paragraph pipeline, we show more examples in Tables \ref{tab:paragraph-translations-chinese} (Chinese), \ref{tab:paragraph-translations-japanese} (Japanese), \ref{tab:paragraph-translations-korean}, and \ref{tab:paragraph-translations-plausible-chinese} (plausible Chinese). The paragraphs were generated with the same sentences in Appendix \ref{sec:sentence-translations}. All shown examples are in the genre ``a news report" (Section \ref{sec:method}), while examples in the other genre can be found in the published data.

\begin{table*}
    \centering
    \tiny
    \begin{tabular}{cl}
        \toprule
        \textbf{Method} & \textbf{Paragraph} \\
        \midrule
        \textsc{Original} & \multicolumn{1}{p{0.8\linewidth}}{\raggedright \chinese{在今日下午的新闻发布会上，被告李先生向媒体披露了一起备受关注的诉讼案件的最新进展。据李先生称，“他们通过威逼利诱，想要我放弃诉讼。”他详细说明了对方如何施加压力，试图迫使其撤销对公司不正当行为的指控。这起事件引起了公众的广泛关注，许多人表示，像李先生这样勇敢站出来揭露真相的行为，值得社会给予更多的支持和保护。}} \\
        \textsc{Baseline} & \multicolumn{1}{p{0.8\linewidth}}{\raggedright At this afternoon's press conference, Mr. Li, the defendant, disclosed the latest developments in a highly publicized lawsuit case to the media. According to Mr. Li, ``They tried to make me give up the lawsuit through threats and bribes." He detailed how the other party applied pressure to force him to withdraw his accusations of the company's misconduct. This incident has garnered widespread public attention, with many people expressing that acts of bravery like Mr. Li's in exposing the truth deserve more societal support and protection.} \\
        \textsc{Faithful Simple} & \multicolumn{1}{p{0.8\linewidth}}{\raggedright At the press conference this afternoon, Mr. Li, the defendant, disclosed to the media the latest developments in a highly publicized lawsuit. According to Mr. Li, ``They tried to make me give up the lawsuit through coercion and bribery." He detailed how the other party applied pressure, attempting to force him to withdraw his allegations of impropriety against the company. This incident has attracted widespread public attention, with many people stating that brave actions like Mr. Li's, in stepping forward to reveal the truth, deserve more societal support and protection.}\\
        \textsc{Creative Simple} & \multicolumn{1}{p{0.8\linewidth}}{\raggedright At this afternoon's press conference, Mr. Li, the defendant, disclosed to the media the latest developments in a highly publicized lawsuit. According to Mr. Li, ``They tried to sway me with a mix of threats and temptations, wanting me to drop the lawsuit." He elaborated on how the other party applied pressure, attempting to force him to withdraw his accusations of improper conduct by the company. This incident has sparked widespread public interest, with many expressing that courageous acts like Mr. Li's in revealing the truth deserve more societal support and protection.}\\
        \textsc{Theme Simple} & \multicolumn{1}{p{0.8\linewidth}}{\raggedright At this afternoon's press conference, Mr. Li, the defendant, revealed to the media the latest developments in a lawsuit that has garnered significant attention. According to Mr. Li, ``They tried to strong-arm and entice me into dropping the lawsuit." He detailed how the other party applied pressure in an attempt to force him to withdraw his accusations of improper conduct by the company. This incident has sparked widespread public interest, with many expressing that courageous acts like Mr. Li's in exposing the truth deserve more societal support and protection.}\\
        \textsc{Faithful CoT} & \multicolumn{1}{p{0.8\linewidth}}{\raggedright At this afternoon's press conference, Mr. Li, the defendant, disclosed to the media the latest developments in a highly watched lawsuit. According to Mr. Li, ``They tried to make me give up the lawsuit through threats and bribes." He detailed how the other party attempted to pressure him into withdrawing his allegations of the company's improper conduct. This incident has garnered widespread public attention, and many people believe that acts of bravery like Mr. Li's, in coming forward to reveal the truth, deserve more societal support and protection.}\\
        \textsc{Creative CoT} & \multicolumn{1}{p{0.8\linewidth}}{\raggedright At this afternoon's press conference, Mr. Li, the defendant, unveiled the latest developments in a highly anticipated lawsuit to the media. Mr. Li claimed, ``They resorted to a mix of intimidation and enticement, aiming for me to drop the lawsuit." He elaborated on how the other party applied pressure, attempting to coerce him into retracting his allegations of misconduct against the company. This incident has drawn widespread public attention, with many voicing that courageous actions like Mr. Li's, stepping forward to expose the truth, deserve greater societal support and protection.}\\
        \textsc{Theme CoT} & \multicolumn{1}{p{0.8\linewidth}}{\raggedright At the press conference this afternoon, Mr. Li, the defendant, disclosed to the media the latest developments in a highly publicized lawsuit. According to Mr. Li, “They tried to make me drop the lawsuit through a combination of threats and inducements.” He detailed how the other party applied pressure, attempting to force him to withdraw his allegations of misconduct against the company. This incident has attracted widespread public attention, with many people expressing that courageous actions like Mr. Li's, in revealing the truth, deserve more support and protection from society.}\\
        \textsc{Faithful Multi-Turn} & \multicolumn{1}{p{0.8\linewidth}}{\raggedright At this afternoon's press conference, Mr. Li, the defendant, disclosed to the media the latest developments in a highly publicized lawsuit. According to Mr. Li, ``They tried to make me give up the lawsuit through coercion and bribery." He detailed how the other party applied pressure, attempting to force him to withdraw his allegations of impropriety against the company. This incident has attracted widespread public attention, with many people stating that brave actions like Mr. Li's, in stepping forward to reveal the truth, deserve more support and protection from society.}\\
        \textsc{Creative Multi-Turn} & \multicolumn{1}{p{0.8\linewidth}}{\raggedright At the news conference this afternoon, Mr. Li, the accused, unveiled to the press the latest developments in a closely watched lawsuit. Mr. Li claimed, ``They attempted to sway me into dropping the lawsuit by a combination of threats and temptations." He elaborated on how the opposition exerted pressure, trying to compel him to rescind his accusations of misconduct against the company. The saga has captivated public interest, with many voicing that courageous disclosures like Mr. Li's merit greater societal endorsement and safeguarding.}\\
        \textsc{Theme Multi-Turn} & \multicolumn{1}{p{0.8\linewidth}}{\raggedright In this afternoon's press conference, Mr. Li, the defendant, shared with the media the recent developments of a lawsuit that has captured widespread attention. Mr. Li stated, ``They resorted to threats and allurements in an effort to make me abandon the lawsuit." He went into detail about how the other side applied pressure, aiming to make him withdraw his claims against the company's wrongful actions. The incident has sparked considerable public interest, with many expressing that the courageous act of coming forward like Mr. Li has done should be met with greater societal support and protection.}\\
        \bottomrule
    \end{tabular}
    \caption{The example paragraph translations for Chinese.}
    \label{tab:paragraph-translations-chinese}
\end{table*}

\begin{table*}
    \centering
    \tiny
    \begin{tabular}{cl}
        \toprule
        \textbf{Method} & \textbf{Paragraph} \\
        \midrule
        \textsc{Original} & \multicolumn{1}{p{0.8\linewidth}}{\raggedright \japanese{日本、東京 - 地元の起業家たちが始めた画期的な取り組みが話題になっています。彼らの開発した環境に優しい再生可能エネルギー製品が予想外の成功を収め、大きな注目を集めています。「私たちの小さなプロジェクトが、まさに燎原之火となり、急速に広がりました。」と、プロジェクトリーダーは興奮を隠せません。この成功は、持続可能な開発と環境保護に関心が集まっている今、他の起業家や企業にも大きな刺激を与えています。}} \\
        \textsc{Baseline} & \multicolumn{1}{p{0.8\linewidth}}{\raggedright Japan, Tokyo - A groundbreaking initiative started by local entrepreneurs is gaining attention. Their development of environmentally friendly renewable energy products has achieved unexpected success and is drawing significant attention. ``Our small project has truly become a blazing wildfire, spreading rapidly," says the project leader, unable to hide their excitement. This success is providing a major stimulus to other entrepreneurs and companies at a time when there is a growing focus on sustainable development and environmental protection.} \\
        \textsc{Faithful Simple} & \multicolumn{1}{p{0.8\linewidth}}{\raggedright Japan, Tokyo - An innovative initiative started by local entrepreneurs is making waves. Their development of eco-friendly renewable energy products has achieved unexpected success and garnered significant attention. ``Our small project has indeed become a wildfire, spreading rapidly," the project leader could hardly contain his excitement. This success is providing a major stimulus to other entrepreneurs and companies at a time when interest in sustainable development and environmental protection is growing.}\\
        \textsc{Creative Simple} & \multicolumn{1}{p{0.8\linewidth}}{\raggedright Tokyo, Japan - A groundbreaking initiative started by local entrepreneurs is becoming a hot topic. Their development of environmentally friendly renewable energy products has achieved unexpected success and has drawn significant attention. ``Our small project has, quite literally, become a wildfire, spreading rapidly," says the project leader, unable to hide his excitement. This success is providing great motivation to other entrepreneurs and corporations at a time when interest in sustainable development and environmental protection is growing.}\\
        \textsc{Theme Simple} & \multicolumn{1}{p{0.8\linewidth}}{\raggedright Japan, Tokyo - An innovative initiative started by local entrepreneurs is becoming a hot topic. Their development of eco-friendly renewable energy products has achieved unexpected success and attracted significant attention. ``Our small project has, indeed, spread like wildfire, rapidly expanding," the project leader said, unable to hide his excitement. This success is providing a substantial stimulus to other entrepreneurs and companies at a time when there is a growing focus on sustainable development and environmental protection.}\\
        \textsc{Faithful CoT} & \multicolumn{1}{p{0.8\linewidth}}{\raggedright Japan, Tokyo - A groundbreaking initiative started by local entrepreneurs is becoming a topic of conversation. Their developed environmentally friendly renewable energy products have achieved unexpected success and are attracting significant attention. ``Our small project has indeed become a raging wildfire, spreading rapidly," says the project leader, unable to hide his excitement. This success is providing a substantial stimulus to other entrepreneurs and companies at a time when there is a strong focus on sustainable development and environmental protection.}\\
        \textsc{Creative CoT} & \multicolumn{1}{p{0.8\linewidth}}{\raggedright Tokyo, Japan - A groundbreaking initiative started by local entrepreneurs is making waves. Their development of environmentally friendly renewable energy products has achieved unexpected success and garnered significant attention. ``Our small project has truly become a wildfire, spreading rapidly," says the project leader, unable to hide their excitement. This success has given a substantial boost to other entrepreneurs and companies at a time when there's a growing focus on sustainable development and environmental protection.}\\
        \textsc{Theme CoT} & \multicolumn{1}{p{0.8\linewidth}}{\raggedright Japan, Tokyo - A groundbreaking initiative started by local entrepreneurs is making headlines. Their development of environmentally friendly renewable energy products has achieved unexpected success and garnered significant attention. ``Our small project has indeed become a wildfire, spreading rapidly," the project leader could not hide their excitement. This success provides a major stimulus to other entrepreneurs and companies at a time when there is growing interest in sustainable development and environmental protection.}\\
        \textsc{Faithful Multi-Turn} & \multicolumn{1}{p{0.8\linewidth}}{\raggedright Japan, Tokyo - An innovative initiative started by local entrepreneurs is gaining attention. Their development of eco-friendly renewable energy products has achieved unexpected success and garnered significant interest. ``Our small project has indeed become a wildfire, spreading rapidly," the project leader could hardly contain his excitement. This success is providing a major stimulus to other entrepreneurs and companies at a time when interest in sustainable development and environmental protection is growing.}\\
        \textsc{Creative Multi-Turn} & \multicolumn{1}{p{0.8\linewidth}}{\raggedright Japan, Tokyo - An innovative initiative started by local entrepreneurs is making waves. Their development of eco-friendly renewable energy products has unexpectedly soared to success, capturing widespread attention. ``Our modest project has, quite literally, turned into a blaze across the prairie, expanding swiftly," the project leader is elated. This triumph is giving a significant boost to other entrepreneurs and corporations at a moment when the spotlight is on sustainable development and conservation of the environment.}\\
        \textsc{Theme Multi-Turn} & \multicolumn{1}{p{0.8\linewidth}}{\raggedright Japan, Tokyo - A groundbreaking initiative launched by local entrepreneurs is creating a buzz. Their invention of eco-friendly renewable energy products has manifested into an unexpected triumph, attracting major interest. ``Our humble project has indeed sparked a conflagration, spreading at an incredible pace," the project leader is visibly thrilled. This success is fueling a significant impetus among other entrepreneurs and businesses at a time when the focus on sustainable development and environmental preservation is intensifying.}\\
        \bottomrule
    \end{tabular}
    \caption{The example paragraph translations for Japanese.}
    \label{tab:paragraph-translations-japanese}
\end{table*}

\begin{table*}
    \centering
    \tiny
    \begin{tabular}{cl}
        \toprule
        \textbf{Method} & \textbf{Paragraph} \\
        \midrule
        \textsc{Original} & \multicolumn{1}{p{0.8\linewidth}}{\raggedright \korean{서울, 대한민국 - 지역 사회에서는 최근 한 개인의 대인 관계 문제가 관심사로 떠올랐습니다. 주변 사람들은 ``진짜 그 사람이 싫어서 안 만나는 건지, 아니면 다른 진짜 이유가 있는 것인지 모르겠어'' 라고 말하며, 이 문제에 대한 의견을 나누고 있습니다. 이 문제는 한 사람의 갑작스런 사회적 활동 감소와 그 이유에 대한 추측으로 인해 생겨났습니다. 해당 개인이나 그의 친구들은 현재 이와 관련된 공식적인 입장을 밝히지 않고 있으며, 이로 인해 더욱 많은 추측이 이어지고 있습니다. 지역사회는 이 문제에 대해 더 많은 정보를 원하며, 이에 대한 진실을 알기 위해 관심을 기울이고 있습니다.}} \\
        \textsc{Baseline} & \multicolumn{1}{p{0.8\linewidth}}{\raggedright Seoul, South Korea - A personal relationship issue of an individual has recently become a matter of interest in the community. People around are sharing their opinions, saying, ``I'm not sure if they're really not meeting because they dislike that person, or if there's another real reason." This issue arose from speculation about the sudden decrease in one person's social activities and the reasons behind it. Neither the individual in question nor their friends have made an official statement regarding the matter, leading to even more speculation. The community is seeking more information on this issue and is keen to find out the truth.} \\
        \textsc{Faithful Simple} & \multicolumn{1}{p{0.8\linewidth}}{\raggedright Seoul, South Korea - In the local community, the personal relationship issue of an individual has recently become a matter of interest. People around are saying, ``I don't know if it's really because they dislike that person and don't want to meet, or if there's another real reason," and are sharing their opinions on this issue. This issue arose due to a sudden decrease in one person's social activities and speculation about the reason. The individual in question or their friends have not made any official statements regarding this matter, leading to further speculation. The community wants more information about this issue and is keen to find out the truth.}\\
        \textsc{Creative Simple} & \multicolumn{1}{p{0.8\linewidth}}{\raggedright Seoul, South Korea - In the local community, the personal relationships of an individual have recently become a matter of interest. People around are saying, ``I wonder if it's really because they dislike that person or if there's another real reason," as they share their opinions on this matter. The issue emerged due to a sudden decrease in social activities by the individual and speculation about the reasons behind it. Neither the individual in question nor their friends have made any official statement regarding this, leading to further speculation. The community is eager for more information on the matter and is paying close attention in hopes of uncovering the truth.}\\
        \textsc{Theme Simple} & \multicolumn{1}{p{0.8\linewidth}}{\raggedright Seoul, South Korea - A personal relationship issue of an individual has recently become a topic of interest in the community. People around are saying, ``I can't tell if they genuinely dislike the person and that's why they're not meeting, or if there's another real reason," and are sharing their opinions on this matter. This issue arose from sudden decreases in one person's social activities and speculation about the reasons behind it. Neither the individual in question nor their friends have made any official statement on the matter, leading to further speculation. The community wants more information on this issue and is keen to uncover the truth.}\\
        \textsc{Faithful CoT} & \multicolumn{1}{p{0.8\linewidth}}{\raggedright Seoul, South Korea - In the local community, the interpersonal relationship issue of an individual has recently arisen as a topic of interest. People around them are saying, “I don't know if they are not meeting them because they really dislike the person, or if there's another real reason,” sharing their opinions on this matter. This issue sprung up due to the sudden decrease in social activities of one person and the speculation about the reasons behind it. The individual in question or their friends have not officially addressed the matter, leading to even more speculation. The community wants more information on this issue and is paying attention to uncover the truth.}\\
        \textsc{Creative CoT} & \multicolumn{1}{p{0.8\linewidth}}{\raggedright Seoul, South Korea - In the local community, the interpersonal issues of an individual have recently become a hot topic. People around are saying, ``I can't tell if they're actually giving the cold shoulder, or if there's some other real reason behind it," sharing their thoughts on the matter. This issue emerged due to a sudden decrease in social activities by an individual and the speculation around the reasons for such withdrawal. Neither the person in question nor their friends have made any official statements regarding the situation, leading to even more speculation. The community is eager for more information and is paying close attention in hopes of uncovering the truth.}\\
        \textsc{Theme CoT} & \multicolumn{1}{p{0.8\linewidth}}{\raggedright Seoul, South Korea - A personal relationship issue of an individual has recently emerged as a concern within the community. People around are saying, ``I don't know whether they're not meeting because they genuinely dislike each other, or if there's some other real reason," sharing their opinions on the matter. This issue arose from speculation about the sudden decrease in social activities of a person and the reasons behind it. Neither the individual in question nor their friends have made any official statement regarding this, leading to even more speculation. The community seeks more information on this issue, paying close attention to uncover the truth.}\\
        \textsc{Faithful Multi-Turn} & \multicolumn{1}{p{0.8\linewidth}}{\raggedright Seoul, South Korea - In the local community, the personal relationship issue of an individual has recently become a matter of interest. People around are saying, ``I don't know if it's really because they dislike the person and don't want to meet, or if there's another real reason," and are sharing their opinions on this issue. This issue arose due to a sudden decrease in one person's social activities and speculation about the reason. The individual in question or their friends have not made any official statements regarding this matter, leading to further speculation. The community wants more information about this issue and is keen to find out the truth.}\\
        \textsc{Creative Multi-Turn} & \multicolumn{1}{p{0.8\linewidth}}{\raggedright Seoul, South Korea - A personal relationship dilemma involving an individual has recently captured the community's attention. Neighbors are speculating, ``Is it genuinely because they despise the person so much they avoid meeting, or is there another underlying truth?" This conversation is buzzing around as people exchange views on the matter. Sparked by one individual's abrupt retreat from social engagements and the swirling guesses about its cause, this topic has become hot. Neither the person at the heart of the rumors nor their close associates have clarified their stance, fueling the rumor mill even more. The community is eager for more clarity on this issue, showing heightened interest to unearth the actual story.}\\
        \textsc{Theme Multi-Turn} & \multicolumn{1}{p{0.8\linewidth}}{\raggedright Seoul, South Korea - In the local community, there's been a rising curiosity about an individual's inter-personal conflict. People around are left wondering, ``Is it true aversion driving them apart or is there something deeper at play?" This speculation has grown from observing a notable drop in one's social interactions, sparking conversations. Neither the individual at the center of these rumors nor their confidants have come forward with any explanations, leading to evolving conjecture. The community is showing a keen interest, hoping to peel back the layers of this mystery and uncover the truth.}\\
        \bottomrule
    \end{tabular}
    \caption{The example paragraph translations for Korean.}
    \label{tab:paragraph-translations-korean}
\end{table*}

\begin{table*}
    \centering
    \tiny
    \begin{tabular}{cl}
        \toprule
        \textbf{Method} & \textbf{Paragraph} \\
        \midrule
        \textsc{Original} & \multicolumn{1}{p{0.8\linewidth}}{\raggedright \chinese{【环球新闻网】随着季节的更迭，大自然总是有着它独特的方式来展现生命之美。在我国南方的一个古老园林中，这种自然的循环得到了完美的体现。每年的春天，园子里春荷盛开，犹如一片片翠绿的玉盘，静静浮在水面上，吸引了无数游人驻足欣赏。而到了秋天，则是秋菊竞艳，黄、紫、白交织的菊花在秋风中摇曳生姿，成为这古园一道别样的风景线。这一切似乎在提醒着人们，无论世事如何变迁，自然的美好总是值得我们去珍惜和维护。}} \\
        \textsc{Baseline} & \multicolumn{1}{p{0.8\linewidth}}{\raggedright [Global News Network] As seasons change, nature always has its unique way of displaying the beauty of life. In an ancient garden in the south of our country, this natural cycle is perfectly embodied. Every spring, the garden is full of blooming spring lotuses, resembling pieces of emerald discs quietly floating on the surface of the water, attracting countless visitors to stop and admire. By autumn, the chrysanthemums outshine each other in beauty; yellow, purple, and white chrysanthemums sway in the autumn breeze, becoming a distinct scenic line in this ancient garden. All this seems to remind us that, no matter how the world changes, the beauty of nature is always worth cherishing and protecting.} \\
        \textsc{Faithful Simple} & \multicolumn{1}{p{0.8\linewidth}}{\raggedright [Global News Network] With the changing of seasons, nature always has its unique way to display the beauty of life. In an ancient garden in the south of our country, this natural cycle is perfectly embodied. Every spring, the garden is full of spring lotus, like pieces of green jade discs, quietly floating on the water surface, attracting countless visitors to stop and admire. And in autumn, it's the time for the autumn chrysanthemums to compete in beauty, with yellow, purple, and white chrysanthemums swaying in the autumn breeze, becoming a different landscape in this ancient garden. All of this seems to remind people that no matter how the world changes, the beauty of nature is always worth our appreciation and preservation.}\\
        \textsc{Creative Simple} & \multicolumn{1}{p{0.8\linewidth}}{\raggedright [Global News Network] As seasons transition, nature always has its unique way of showcasing the beauty of life. In an ancient garden in the southern part of our country, this natural cycle is perfectly embodied. Each spring, the garden blooms with lotus, like emerald discs quietly floating on the water, attracting countless visitors to stop and admire. Come autumn, it's a competition of chrysanthemums' beauty, with yellow, purple, and white chrysanthemums swaying in the autumn breeze, forming a distinctive landscape in this ancient garden. All these seem to remind us that, no matter how the world changes, the beauty of nature is always worth cherishing and protecting.}\\
        \textsc{Theme Simple} & \multicolumn{1}{p{0.8\linewidth}}{\raggedright [Global News Network] With the change of seasons, Nature always has its unique way of showcasing the beauty of life. In an ancient garden in the southern part of our country, this natural cycle is perfectly represented. Every spring, the garden is full of blooming lotus, like pieces of emerald green jade plates quietly floating on the water, attracting countless visitors to stop and admire. And when autumn comes, it's time for the chrysanthemums to compete in beauty, with yellow, purple, and white chrysanthemums swaying in the autumn breeze, becoming a unique landscape of this ancient garden. All this seems to remind us that no matter how the world changes, the beauty of nature is always worth cherishing and protecting.}\\
        \textsc{Faithful CoT} & \multicolumn{1}{p{0.8\linewidth}}{\raggedright [Global News Network] As the seasons change, nature always has its unique way of showcasing the beauty of life. This natural cycle is perfectly embodied in an ancient garden in the southern part of our country. Every spring, the garden is filled with spring lotuses, which bloom like green jade discs quietly floating on the water, attracting countless visitors to stop and admire. When autumn arrives, it's the autumn chrysanthemums that compete in beauty, with yellow, purple, and white chrysanthemums swaying in the autumn breeze, forming a distinct landscape in this ancient garden. All of this seems to remind people that no matter how the world changes, the beauty of nature is always worth our appreciation and preservation.}\\
        \textsc{Creative CoT} & \multicolumn{1}{p{0.8\linewidth}}{\raggedright **Global News Network** - As seasons change, nature always finds its unique way to display the beauty of life. In an ancient garden in the south of our country, this natural cycle is perfectly exemplified. Every spring, the garden is adorned with lotus blossoms, like myriad verdant jade discs quietly floating on the water, drawing countless visitors to pause and admire. Come autumn, it's the chrysanthemums' turn to shine, with yellow, purple, and white blooms dancing in the autumn breeze, creating a distinct scenic line in this ancient garden. All these seem to remind us that, no matter how the world changes, the splendor of nature is always worth our appreciation and preservation.}\\
        \textsc{Theme CoT} & \multicolumn{1}{p{0.8\linewidth}}{\raggedright [Global News Network] As the seasons change, nature always has its unique way of showcasing the beauty of life. In an ancient garden in the southern part of our country, this natural cycle is perfectly embodied. In spring, the garden is filled with blooming lotuses, like green jade plates quietly floating on the water, attracting countless visitors to stop and admire. In autumn, it's the chrysanthemums' turn to dazzle, with yellow, purple, and white blooms swaying in the autumn breeze, presenting a different kind of scenery in the ancient garden. All of this seems to remind us that no matter how the world changes, the beauty of nature is always worth cherishing and protecting.}\\
        \textsc{Faithful Multi-Turn} & \multicolumn{1}{p{0.8\linewidth}}{\raggedright [Global News Network] As the seasons change, nature always has its unique way of showcasing the beauty of life. In an ancient garden in the south of our country, this natural cycle is perfectly embodied. Each spring, the garden blooms with spring lotus, like pieces of emerald green jade discs, quietly floating on the water's surface, attracting countless visitors to stop and admire. In autumn, it's the turn of the autumn chrysanthemums to compete in beauty, with yellow, purple, and white chrysanthemums swaying in the autumn breeze, becoming a distinctive landscape in this ancient garden. All of this seems to remind us that no matter how the world changes, the beauty of nature is always worth cherishing and preserving.}\\
        \textsc{Creative Multi-Turn} & \multicolumn{1}{p{0.8\linewidth}}{\raggedright [Global News Network] With the seasonal shifts, nature unfurls the splendor of life in its unique expressions. Within the verdant confines of an ancient garden in our country's south, this rhythmic cadence of nature is flawlessly manifested. Come spring, the garden is a spectacle of spring lotuses, mirroring a collection of emerald platters, serenely adrift upon the waters, luring myriad visitors to pause and revel in their beauty. Autumn heralds the reign of chrysanthemums, weaving a tapestry of yellow, purple, and white against the autumnal air, presenting a contrasting spectacle to the observer. Through this endless cycle of spring lotus and autumn chrysanthemum, nature subtly reminds us - amidst the ever-changing tapestry of life, its inherent beauty remains a constant, beckoning our appreciation and preservation.}\\
        \textsc{Theme Multi-Turn} & \multicolumn{1}{p{0.8\linewidth}}{\raggedright [Global News Network] As the calendar pages turn, nature invariably reveals the splendor of existence in its distinct manner. In a venerable garden in the south of our nation, this perpetual dance of nature is exquisitely showcased. When spring arrives, the garden is awash with spring lotuses, resembling emerald plates, serenely floating on the water, enticing legions of visitors to halt and admire. Come autumn, the garden transitions to showcase the rival beauty of autumn chrysanthemums, with blooms of yellow, purple, and white swaying in the autumn breeze, crafting an alternative picturesque scene in this ancient park. This eternal sequence of lotus blossoms in spring and chrysanthemums in fall gently nudges us to remember that, despite the shifts and turns of worldly affairs, the beauty encapsulated within nature is always there, worthy of our protection and esteem.}\\
        \bottomrule
    \end{tabular}
    \caption{The example paragraph translations for plausible Chinese.}
    \label{tab:paragraph-translations-plausible-chinese}
\end{table*}

\section{Scores}
\label{sec:scores}

\begin{figure*}
    \centering
    \includegraphics[width=\linewidth]{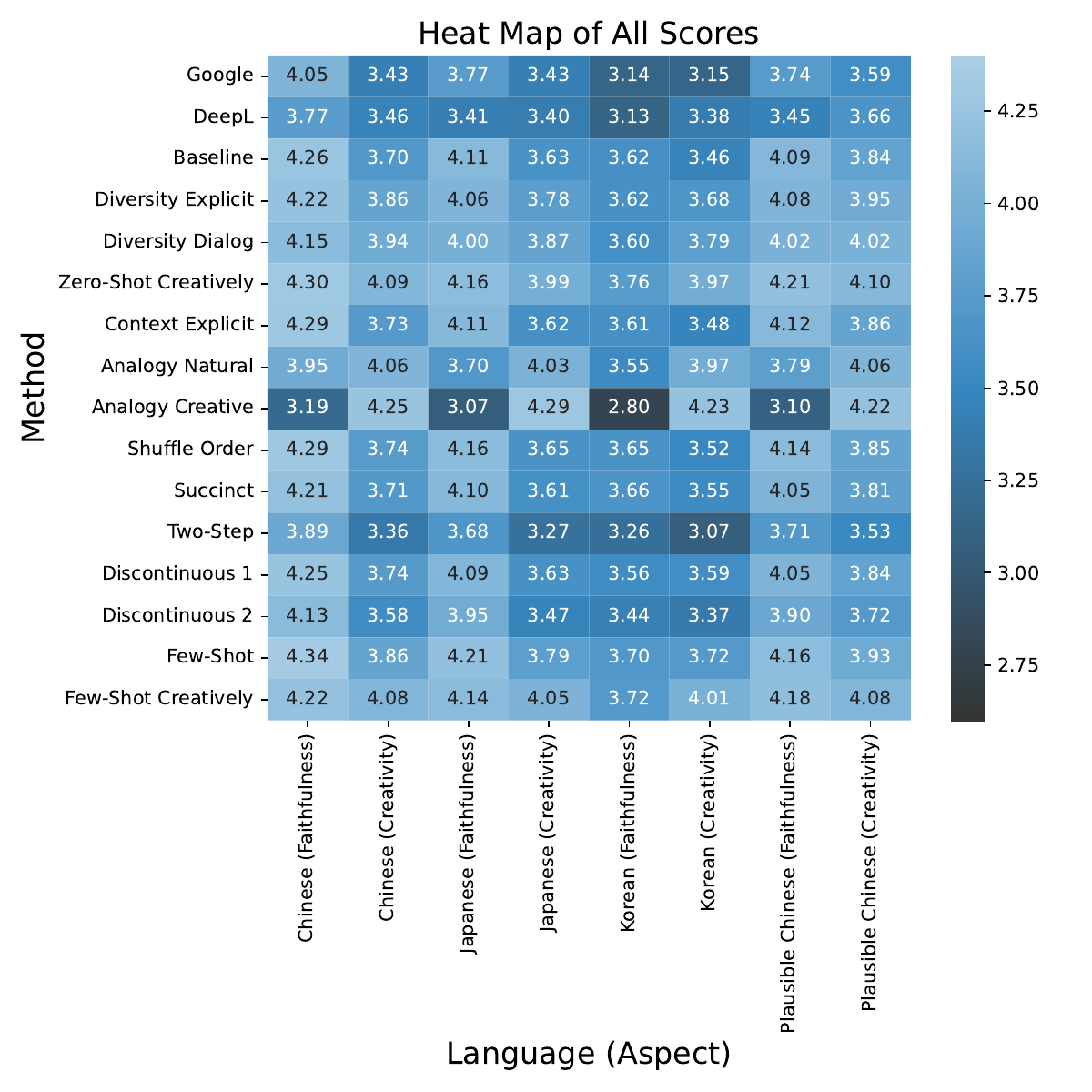}
    \caption{The heat map for all faithfulness and creativity scores.}
    \label{fig:all-scores-heat-map}
\end{figure*}

\begin{figure*}
    \centering
    \includegraphics[width=\columnwidth]{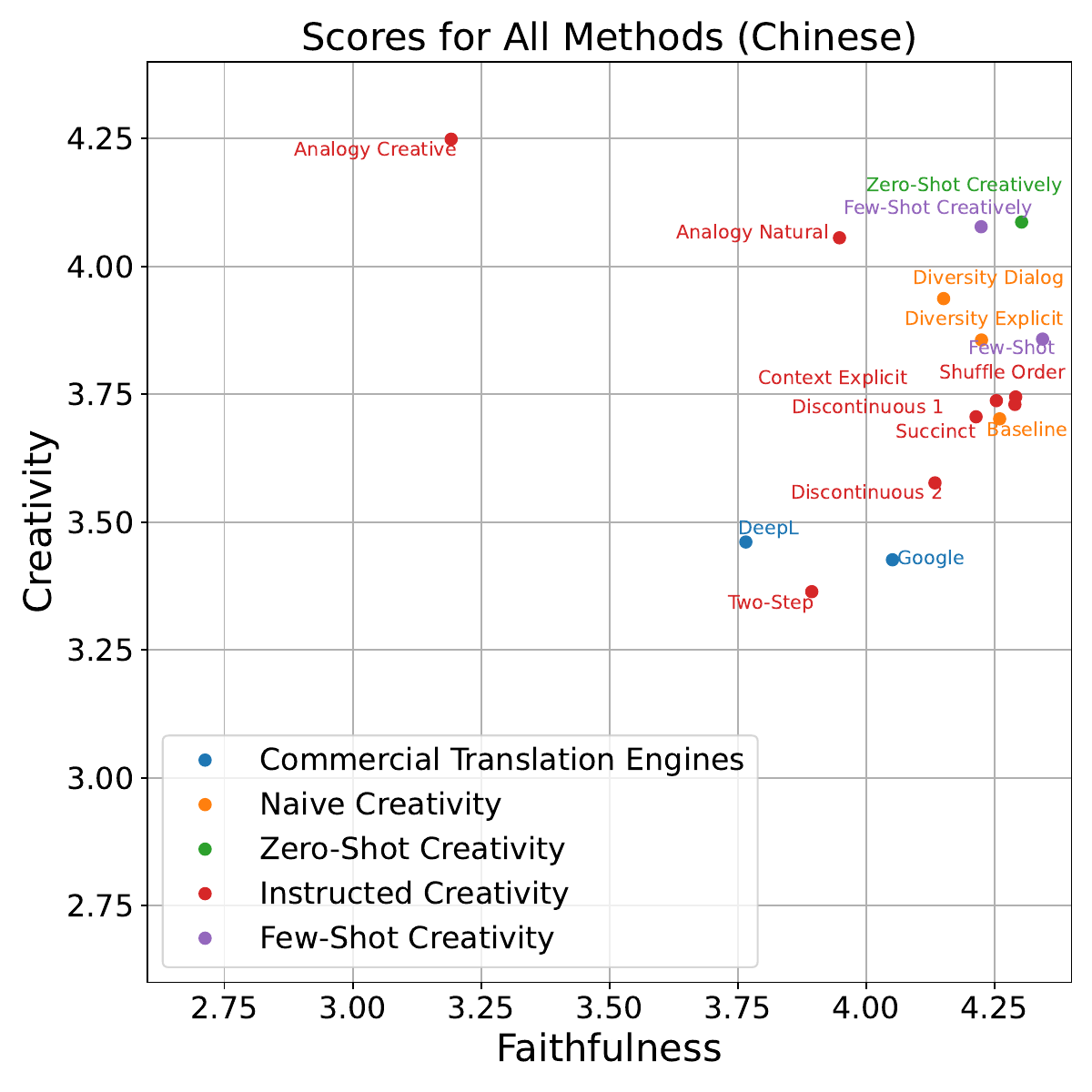}
    \includegraphics[width=\columnwidth]{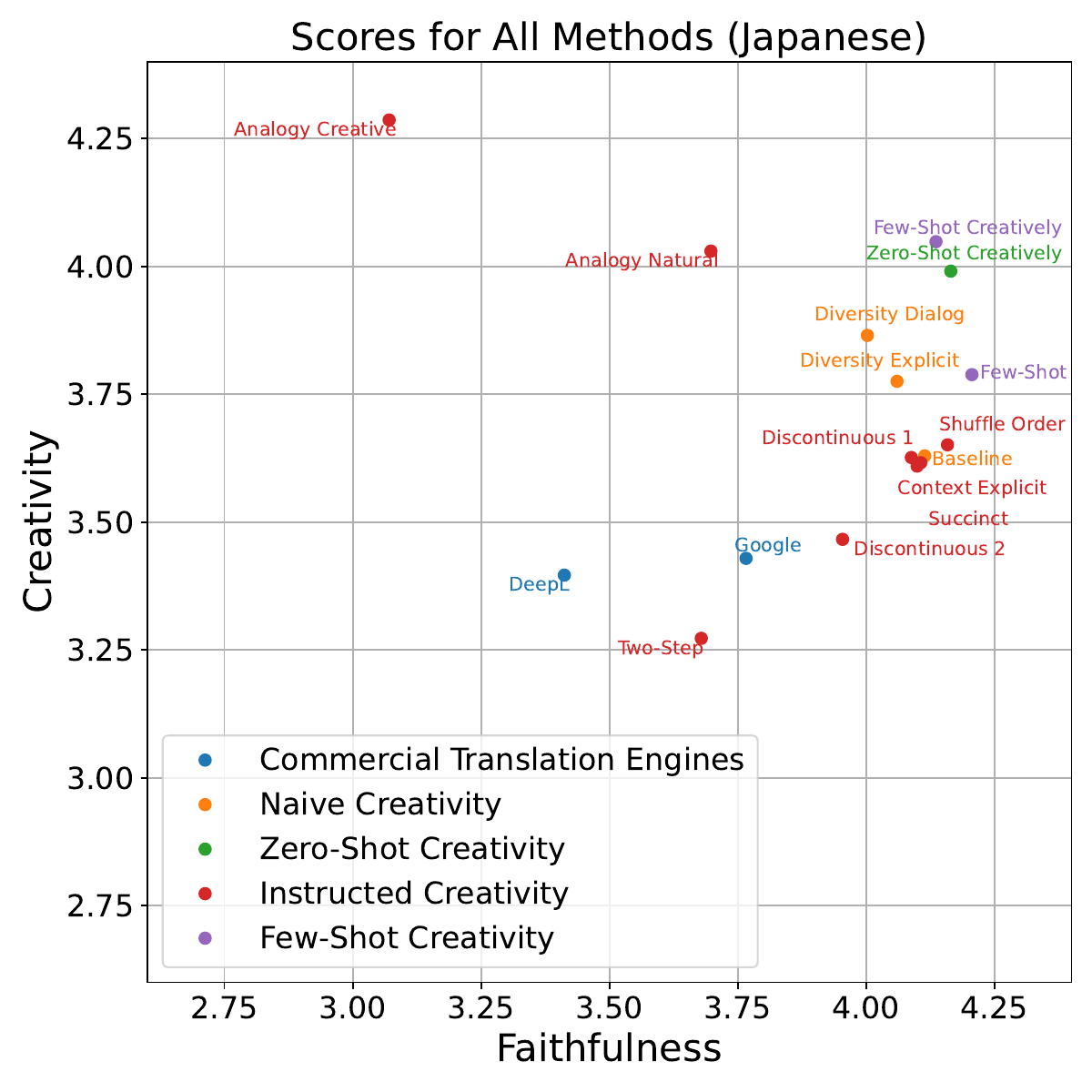}
    \includegraphics[width=\columnwidth]{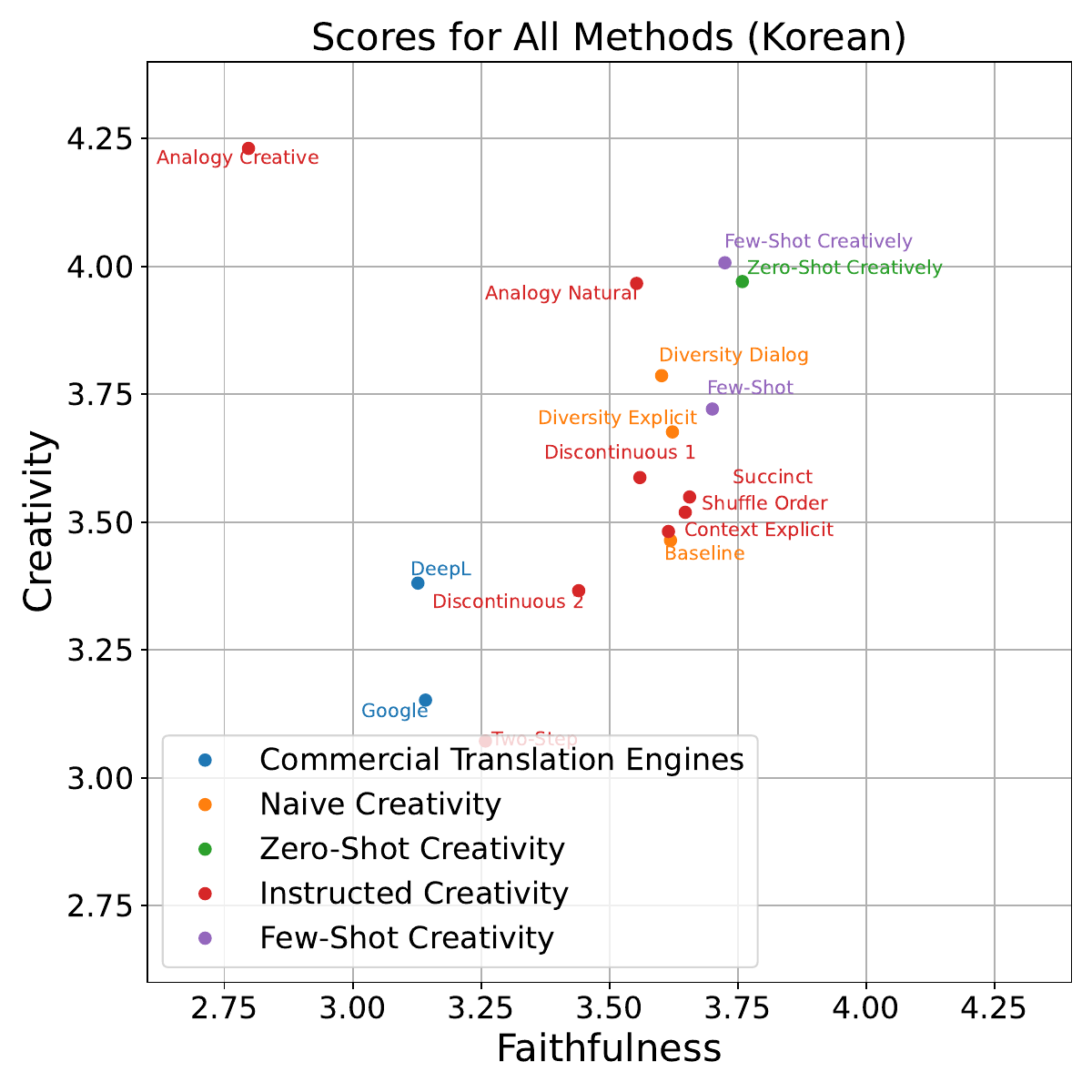}
    \includegraphics[width=\columnwidth]{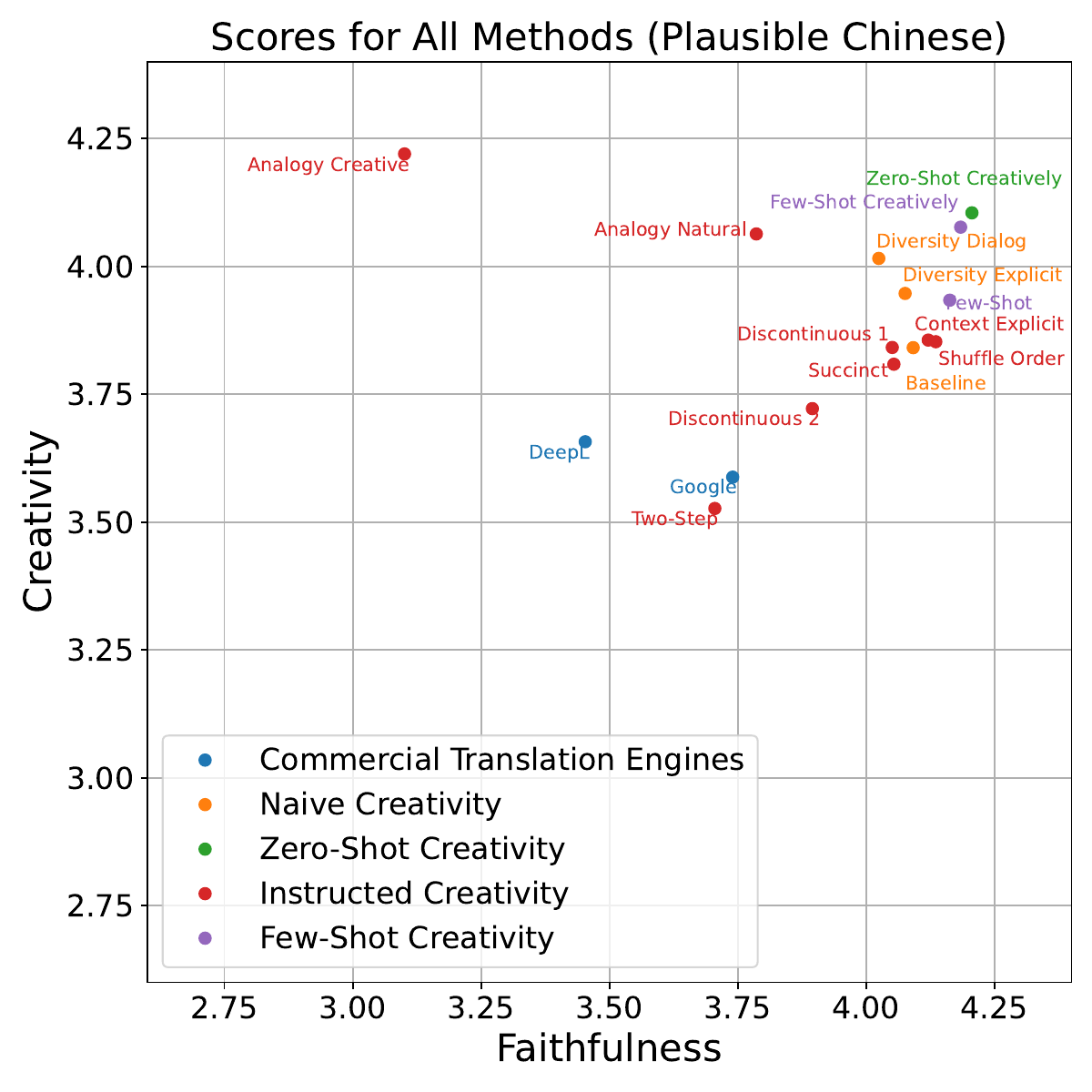}
    \caption{The scatter plots for all faithfulness and creativity scores.}
    \label{fig:all-scores-scatter}
\end{figure*}

We summarize the faithfulness and creativity scores for all languages and all strategies in Table \ref{tab:all-scores}. We also visualize the scores using a heat map in Figure \ref{fig:all-scores-heat-map} and scatter plots in Figure \ref{fig:all-scores-scatter}.

\begin{table*}
\centering
\tiny
\begin{tabular*}{\textwidth}{@{\extracolsep{\fill}}*{10}{c}}
\toprule
\multirow{2}{*}{\textbf{Method}} & \multirow{2}{*}{\textbf{Size}} & \multicolumn{2}{c}{\textbf{Chinese}} & \multicolumn{2}{c}{\textbf{Japanese}} & \multicolumn{2}{c}{\textbf{Korean}} & \multicolumn{2}{c}{\textbf{Plausible Chinese}} \\
\cmidrule{3-4}\cmidrule{5-6}\cmidrule{7-8}\cmidrule{9-10}
                 &  & \textbf{Faithfulness} & \textbf{Creativity} & \textbf{Faithfulness} & \textbf{Creativity} & \textbf{Faithfulness} & \textbf{Creativity} & \textbf{Faithfulness} & \textbf{Creativity}\\
\midrule
\textsc{Google} &  500 & 4.05 $\pm$ 0.76 & 3.43 $\pm$ 0.52 & 3.77 $\pm$ 0.96 & 3.43 $\pm$ 0.56 & 3.14 $\pm$ 1.00 & 3.15 $\pm$ 0.64 & 3.74 $\pm$ 0.86 & 3.59 $\pm$ 0.48 \\
\textsc{DeepL} &  500 & 3.77 $\pm$ 1.00 & 3.46 $\pm$ 0.58 & 3.41 $\pm$ 1.06 & 3.40 $\pm$ 0.58 & 3.13 $\pm$ 1.01 & 3.38 $\pm$ 0.60 & 3.45 $\pm$ 1.00 & 3.66 $\pm$ 0.48 \\
\textsc{Baseline} &  500 & 4.26 $\pm$ 0.59 & 3.70 $\pm$ 0.43 & 4.11 $\pm$ 0.73 & 3.63 $\pm$ 0.49 & 3.62 $\pm$ 0.83 & 3.46 $\pm$ 0.52 & 4.09 $\pm$ 0.67 & 3.84 $\pm$ 0.37 \\
\textsc{Diversity Explicit} & 2500 & 4.22 $\pm$ 0.53 & 3.86 $\pm$ 0.33 & 4.06 $\pm$ 0.63 & 3.78 $\pm$ 0.39 & 3.62 $\pm$ 0.77 & 3.68 $\pm$ 0.45 & 4.08 $\pm$ 0.57 & 3.95 $\pm$ 0.29 \\
\textsc{Diversity Dialog} & 2500 & 4.15 $\pm$ 0.51 & 3.94 $\pm$ 0.27 & 4.00 $\pm$ 0.59 & 3.87 $\pm$ 0.34 & 3.60 $\pm$ 0.75 & 3.79 $\pm$ 0.39 & 4.02 $\pm$ 0.55 & 4.02 $\pm$ 0.25 \\
\textsc{Zero-Shot Creatively} &  500 & 4.30 $\pm$ 0.50 & 4.09 $\pm$ 0.23 & 4.16 $\pm$ 0.58 & 3.99 $\pm$ 0.27 & 3.76 $\pm$ 0.68 & 3.97 $\pm$ 0.33 & 4.21 $\pm$ 0.55 & 4.10 $\pm$ 0.28 \\
\textsc{Context Explicit} & 2000 & 4.29 $\pm$ 0.58 & 3.73 $\pm$ 0.42 & 4.11 $\pm$ 0.70 & 3.62 $\pm$ 0.49 & 3.61 $\pm$ 0.81 & 3.48 $\pm$ 0.55 & 4.12 $\pm$ 0.65 & 3.86 $\pm$ 0.36 \\
\textsc{Analogy Natural} &  500 & 3.95 $\pm$ 0.70 & 4.06 $\pm$ 0.27 & 3.70 $\pm$ 0.74 & 4.03 $\pm$ 0.27 & 3.55 $\pm$ 0.72 & 3.97 $\pm$ 0.28 & 3.79 $\pm$ 0.78 & 4.06 $\pm$ 0.25 \\
\textsc{Analogy Creative} &  500 & 3.19 $\pm$ 0.91 & 4.25 $\pm$ 0.31 & 3.07 $\pm$ 0.88 & 4.29 $\pm$ 0.32 & 2.80 $\pm$ 0.91 & 4.23 $\pm$ 0.35 & 3.10 $\pm$ 0.91 & 4.22 $\pm$ 0.29 \\
\textsc{Shuffle Order} &  500 & 4.29 $\pm$ 0.56 & 3.74 $\pm$ 0.40 & 4.16 $\pm$ 0.63 & 3.65 $\pm$ 0.47 & 3.65 $\pm$ 0.77 & 3.52 $\pm$ 0.49 & 4.14 $\pm$ 0.63 & 3.85 $\pm$ 0.36 \\
\textsc{Succinct} &  500 & 4.21 $\pm$ 0.61 & 3.71 $\pm$ 0.44 & 4.10 $\pm$ 0.67 & 3.61 $\pm$ 0.49 & 3.66 $\pm$ 0.77 & 3.55 $\pm$ 0.48 & 4.05 $\pm$ 0.68 & 3.81 $\pm$ 0.38 \\
\textsc{Two-Step} &  500 & 3.89 $\pm$ 0.70 & 3.36 $\pm$ 0.57 & 3.68 $\pm$ 0.85 & 3.27 $\pm$ 0.57 & 3.26 $\pm$ 0.90 & 3.07 $\pm$ 0.61 & 3.71 $\pm$ 0.82 & 3.53 $\pm$ 0.52 \\
\textsc{Discontinuous 1} &  500 & 4.25 $\pm$ 0.60 & 3.74 $\pm$ 0.39 & 4.09 $\pm$ 0.67 & 3.63 $\pm$ 0.47 & 3.56 $\pm$ 0.85 & 3.59 $\pm$ 0.50 & 4.05 $\pm$ 0.68 & 3.84 $\pm$ 0.34 \\
\textsc{Discontinuous 2} &  500 & 4.13 $\pm$ 0.65 & 3.58 $\pm$ 0.49 & 3.95 $\pm$ 0.78 & 3.47 $\pm$ 0.53 & 3.44 $\pm$ 0.89 & 3.37 $\pm$ 0.60 & 3.90 $\pm$ 0.80 & 3.72 $\pm$ 0.44 \\
\textsc{Few-Shot} &  500 & 4.34 $\pm$ 0.53 & 3.86 $\pm$ 0.33 & 4.21 $\pm$ 0.65 & 3.79 $\pm$ 0.40 & 3.70 $\pm$ 0.77 & 3.72 $\pm$ 0.44 & 4.16 $\pm$ 0.59 & 3.93 $\pm$ 0.33 \\
\textsc{Few-Shot Creatively} &  500 & 4.22 $\pm$ 0.61 & 4.08 $\pm$ 0.27 & 4.14 $\pm$ 0.64 & 4.05 $\pm$ 0.36 & 3.72 $\pm$ 0.71 & 4.01 $\pm$ 0.37 & 4.18 $\pm$ 0.56 & 4.08 $\pm$ 0.28 \\
\bottomrule
\end{tabular*}
\caption{The faithfulness and creativity scores for all strategies and all languages.}
\label{tab:all-scores}
\end{table*}

\end{document}